%% file: main.tex
\documentclass[10pt,twocolumn,letterpaper]{article}

\usepackage[pagenumbers]{cvpr} % To force page numbers, e.g. for an arXiv version


\definecolor{cvprblue}{rgb}{0.21,0.49,0.74}
\usepackage[pagebackref,breaklinks,colorlinks,allcolors=cvprblue]{hyperref}

\usepackage{soul}
\usepackage{float}

\usepackage{multirow}
\usepackage{color, colortbl} % For table colors
\usepackage{pifont}
\newcommand{\cmark}{$\checkmark$}%
\newcommand{\xmark}{$\times$}%
\usepackage{caption}
\usepackage{tabularray}
\usepackage{algorithm}
\usepackage{algorithmic}

\def\paperID{13517} % *** Enter the Paper ID here
\def\confName{CVPR}
\def\confYear{2025}

\newcommand{\sota}{state-of-the-art\xspace}

\newcommand{\mm}{Markov-map\xspace}
\newcommand{\mms}{Markov-maps\xspace}

\definecolor{highlight}{rgb}{0.95,0.95,1}
\title{Repurposing Stable Diffusion Attention for Training-Free Unsupervised Interactive Segmentation}
\author{Markus Karmann\ \ \ \ \ \ \ \ \ Onay Urfalioglu\\
\tt{Vivo Tech Research GmbH}}

\begin{document}

\makeatletter
\g@addto@macro\@maketitle{
  \begin{figure}[H]
  \setlength{\linewidth}{\textwidth}
  \setlength{\hsize}{\textwidth}
  \centering
  \includegraphics[width=\textwidth]{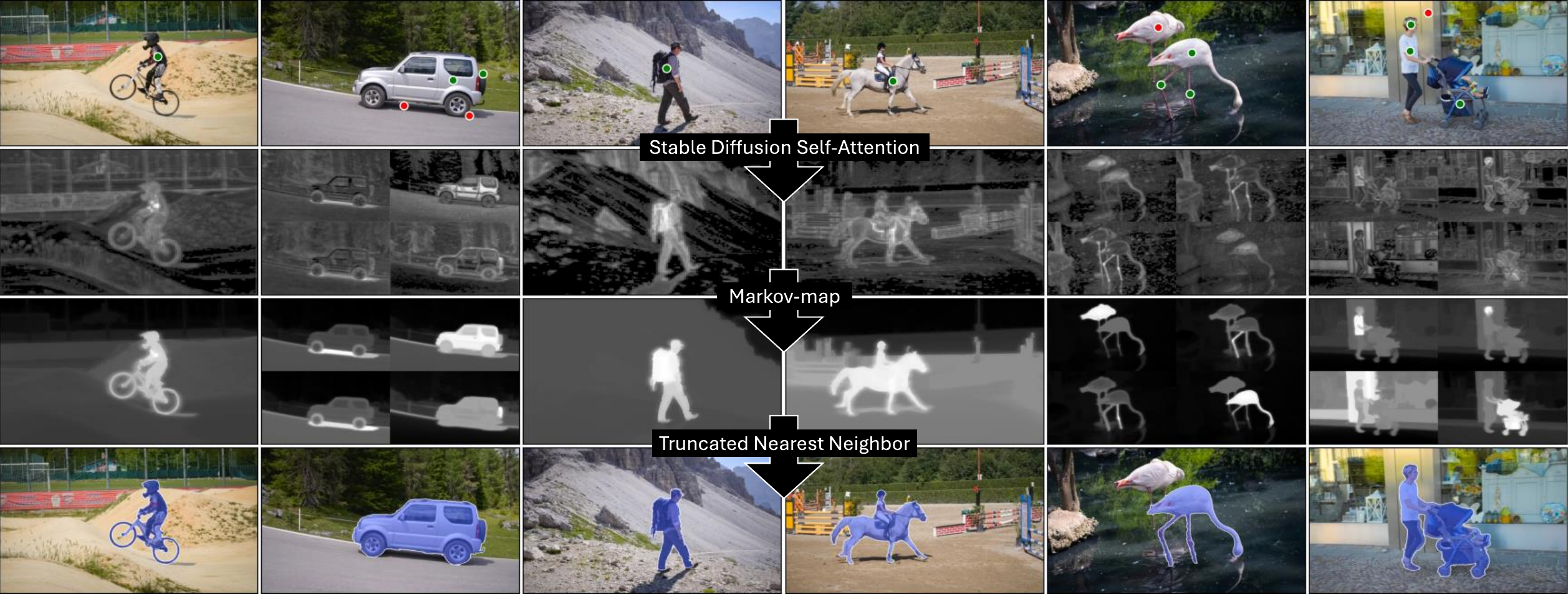}
  \caption{\textbf{We introduce M2N2, an unsupervised training-free point prompt based segmentation framework.} We enhance the semantic information present in the self-attention of Stable Diffusion 2 by using a Markov process to generate semantically rich Markov-maps. We then perform a truncated nearest neighbor of each point's Markov-map to obtain a final segmentation.}
  \end{figure}
}
\makeatother

\maketitle
\input{sec/0_abstract}  
\input{sec/1_intro}
\input{sec/related_work}
\input{sec/3_method}

\input{sec/4_experiments}
\input{sec/conclusions}
\clearpage
{
    \small
    \bibliographystyle{ieeenat_fullname}
    \newcommand{\BIBdecl}{\setlength{\itemsep}{0.25 em}}
    %\raggedright
    \bibliography{main}
}

% WARNING: do not forget to delete the supplementary pages from your submission 
\input{sec/X_suppl}

\end{document}

%% file: sec/0_abstract.tex
%\begin{figure*}[t]
%    \centering
%    \includegraphics[width=\textwidth]{figures/Thumbnail.pdf}
%    \caption{Just out of curiousty, is the flow of the text working? \todo{yes, this has to be changed to an actual describtion}}
%    \label{fig:thumbnail}
%\end{figure*}

\begin{abstract}
Recent progress in interactive point prompt based Image Segmentation allows to significantly reduce the manual effort to obtain high quality semantic labels.
State-of-the-art unsupervised methods use self-supervised pre-trained models to obtain pseudo-labels which are used in training a prompt-based segmentation model.
In this paper, we propose a novel unsupervised and training-free approach based solely on the self-attention of Stable Diffusion.
We interpret the self-attention tensor as a Markov transition operator, which enables us to iteratively construct a Markov chain.
Pixel-wise counting of the required number of iterations along the Markov chain to reach a relative probability threshold yields a Markov-iteration-map, which we simply call a Markov-map.
Compared to the raw attention maps, we show that our proposed Markov-map has less noise, sharper semantic boundaries and more uniform values within semantically similar regions.
We integrate the Markov-map in a simple yet effective truncated nearest neighbor framework to obtain interactive point prompt based segmentation.
Despite being training-free, we experimentally show that our approach yields excellent results in terms of Number of Clicks (NoC), even outperforming 
state-of-the-art training based unsupervised methods in most of the datasets.
Code is available at \url{https://github.com/mkarmann/m2n2}.
\end{abstract}

%% file: sec/1_intro.tex
\section{Introduction}
\label{sec:intro}
The goal of point prompt based interactive image segmentation is to obtain a high-quality segmentation from limited user interaction in the form of clicking.
Prompt-based image segmentation gained popularity lately due to large supervised foundation models~\cite{SAM, SAM2}.
In this paper, we focus on unsupervised methods, where no segmentation labels are used at all in the design and/or training of the models.
Most recent approaches rely on self supervised backbones like ViT~\cite{vit}, trained either by DINO~\cite{DINO} or MAE~\cite{MAE} based self-supervised techniques.
On the other hand, Stable Diffusion (SD)~\cite{StableDiffusion} has been used for many different computer vision applications such as monocular depth estimation~\cite{Repurposing_SD_for_MonoDepth}, semantic segmentation~\cite{DiffuseAttendSegment}, object detection~\cite{DiffusionDet} and image classification~\cite{SD_Secretly_Zero_Shot_Classifier}, often resulting in state-of-the-art solutions. 

Inspired by DiffSeg~\cite{DiffuseAttendSegment}, we investigate the potential of SD's self-attention maps for training-free interactive segmentation.
Generally, for all self-supervised backbones, the main challenges are that attention maps don't distinguish between instances, exhibit noise and are typically of low resolution.
We choose SD, particularly version 2 (SD2), as the backbone of our approach, as it provides the highest attention resolution (\cref{sec:choice_of_backbone}).
To overcome the mentioned challenges, we interpret the self-attention tensors as a Markov transition operator, where the repeated application of the transition forms a Markov chain.
We propose a novel Markov-iteration-map or simply Markov-map, where each pixel counts the number of iterations required to obtain a specific probability value.
We show that the proposed Markov-map has less noise, the semantic boundaries are sharper and the semantic regions within Markov-maps are more uniformly distributed.
Native Markov-maps do not distinguish between instances.
Therefore, we further improve Markov-maps with a flood fill approach, which suppresses local minima, to enable instance based segmentation.
Finally, we obtain Markov-maps of each prompt point and combine them with a truncated nearest neighbor approach to enable multi-prompt point interactive segmentation.
Surprisingly, despite being training-free, we significantly improve the state-of-the-art in terms of Number of Clicks (NoC) and even surpass training based unsupervised approaches in most of the datasets.
Our main contributions are:
\begin{itemize}
    \item We introduce {\bf M}arkov-{\bf M}ap {\bf N}earest {\bf N}eighbor ({\bf M2N2}), the first attention-based unsupervised training-free multi-point prompt segmentation framework.
    \item We propose a novel method to refine semantic information in attention maps, which we call {\bf Markov-maps}.
    \item We enable instance aware Markov-maps by utilizing a modified flood fill approach.
    \item We introduce a truncated nearest neighbor approach to combine multiple point prompts.
    \item We conduct extensive experiments with multiple backbones and achieve \sota results, surpassing even unsupervised training-based methods.
\end{itemize}
% The remaining of the paper is organized as follows: in~\cref{sec:related_work}
% we discuss related work followed by~\cref{sec:method}, where we 
% introduce our proposed method in full detail. In~\cref{sec:experiments}, we 
% present experimental results and in~\cref{sec:conclusions}, we conclude our paper.

\begin{figure*}[!ht]
  \centering
  \includegraphics[width=\textwidth]{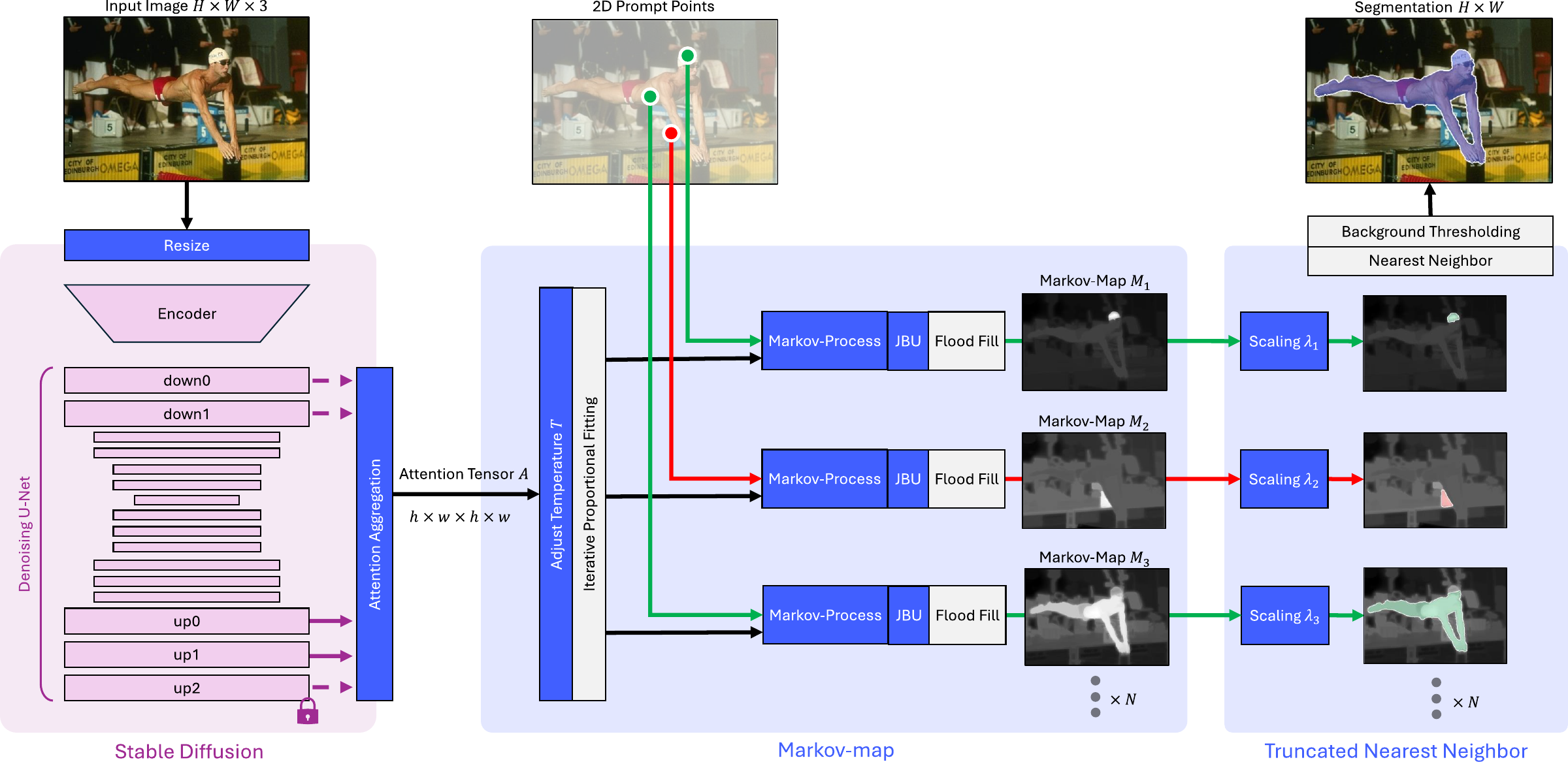}
  \caption{\textbf{M2N2 framework overview}. We perform a single denoising step of the input image with Stable Diffusion 2 to obtain attention tensors. The tensors are aggregated and utilized to obtain a \mm $M_i$ for each prompt point. The final segmentation is the result of a truncated nearest neighbor of scaled \mms $M_i$ as a measure of semantic distance for each prompt point. The green and red areas in the scaled \mms denote regions where the distance is less or equal to the global background threshold. In this visualization, components in blue contain adjustable hyperparameters.}
  \label{fig:framework_overview}
\end{figure*}

%% file: sec/related_work.tex
\section{Related Work}\label{sec:related_work}
Point prompt based interactive image segmentation has been approached from multiple perspectives. In this paper, we distinguish between supervised and unsupervised methods.
\subsection{Supervised Methods}
%Supervised methods can be grouped by dense fusion and sparse fusion of prompt information. Dense prompt maps are utilized in ~\cite{DIOS}

In~\cite{DIOS} a click map and click sampling strategies are used in combination with FCN~\cite{FCNSemanticSegmentation}, creating the foundation for many follow-up methods.
More recent approaches include ~\cite{ritm2022}, ~\cite{interactiveSegmentationFirstClickAttn},~\cite{SAM, HQSAM}, ~\cite{Rana:Dynamite}, ~\cite{interactiveSegmentationbackpropRefinement, interactiveSegmentationfBRS}, ~\cite{SimpleClick}, 
~\cite{interactiveSegmentationGaussianProcess}, ~\cite{interactiveSegmentationFocusCut, FocalClick}, 
~\cite{clickPromptOptimalTransport}, ~\cite{CFR-ICL}, ~\cite{SegNext-InteractiveSegmentation},~\cite{interformer}, ~\cite{OISegmentation}, either 
deploying dense fusion, where prompts are encoded as prompt maps, or sparse fusion, where prompts are transformed into embedding space to be fused with image and other related embeddings.

Although the supervised approaches achieve good performance and efficiency, they require large-scale pixel-level annotations to train, which are expensive and laborious to obtain. 
While many supervised methods are tested on additional domains like medical images, it is unclear if there are other domains where the trained models would have a domain gap.
%Most of the recent \sota approaches are based on some large model as a backbone to extract features from. Typically, such backbone models are trained on a large number of images and provide rich features which can be used to create pseudo labels for semantic segmentation. 

\subsection{Unsupervised methods}
Classical, unsupervised methods not based on Deep Learning like GraphCut~\cite{grabcut}, Random Walk~\cite{randomWalkSegmentation}, Geodesic Matting~\cite{geodesicMattingSegmentation}, GSC~\cite{geodesicStartSegmentation} and ESC~\cite{geodesicStartSegmentation} have been proposed.
However, recently Deep Learning based approaches show great potential. Especially, methods utilizing self-supervised learning achieve impressive results.
Such methods rely on pre-trained models (e.g., DenseCL~\cite{DenseCL}, DINO~\cite{DINO}) to extract segments from their features. 
In~\cite{Simoni2021LocalizingOW} some heuristics are proposed to choose pixels belonging to the same object according to
their feature similarity. 
\cite{Wang2022SelfSupervisedTF} introduces normalized cuts~\cite{Shi00normalizedCuts} on the affinity graph constructed by pixel-level representations from DINO to divide the foreground and background of an image. 
In~\cite{hamilton2022unsupervised} a segmentation head is trained by distilling the feature correspondences from DINO. 
\cite{DSM} adopts spectral decomposition on the affinity graph to discover meaningful parts in an image. 
FreeSOLO~\cite{FreeSOLO} designs pseudo instance mask generation based on multi-scale feature correspondences from densely pre-trained models and trains an instance segmentation model with these pseudo masks. \\
Recently, several papers used Stable Diffusion (SD)~\cite{StableDiffusion} for various kinds of applications targeting unsupervised Semantic Segmentation. 
In~\cite{DiffuseAttendSegment} the self-attention maps of SD were used with KL-Divergence based similarity measures to merge semantically similar regions in order to extract segments.
MIS~\cite{MIS} and UnSAM~\cite{wang2024segment} use methods to capture the semantic hierarchy and create pseudo labels, which are used in a follow-up training process to obtain a model for promptable segmentation. UnSAM additionally allows automatic whole-image segmentation, but does not report NoC values on any dataset.
In~\cite{DiffusionDet} object detection is formulated as a denoising diffusion process from noisy boxes to object boxes, and~\cite{paintSeg} proposes training-free unsupervised segmentation using a pre-trained diffusion model by iteratively contrasting the original image and a painted image in which a masked area is re-painted. 
While this method obtains good results, it does not provide multi-prompt interactive segmentation.

Inspired by~\cite{DiffuseAttendSegment}, we focus on SD's self-attention maps as an initial semantic feature map.  
Our proposed Markov-map improves the semantic features, resulting in an excellent, unsupervised point prompt based interactive segmentation solution. 

%% file: sec/3_method.tex
\section{Method}
\label{sec:method}
In this section, we introduce our \textbf{M}arkov-\textbf{M}ap \textbf{N}earest \textbf{N}eighbor (M2N2) framework in full detail.
We design our interactive process similar to \cite{SimpleClick}\cite{SAM} where the user is shown an image with a suggested (initially empty) segmentation and allowing them to place a foreground or background point on objects they want to either include or exclude from the segmentation.
Once a new point is set, M2N2 uses the entire set of points to predict a new segmentation.
This interactive process continues until a satisfactory segmentation is achieved.
\cref{fig:framework_overview} shows the three main stages of M2N2 applied on an example in which three points have already been set.
In the first stage, we obtain an attention tensor of the input image by aggregating SD's self-attentions.
The second stage extracts and enhances the semantic information of the attention tensor by creating a \mm for each prompt point.
The semantically rich \mms are then utilized in a truncated nearest neighbor algorithm to obtain a training-free unsupervised segmentation.
In the following subsections, we first formulate the problem in the context of a nearest neighbor algorithm in \cref{sec:truncated_nn}, followed by an explanation of the acquisition of the attention tensor in \cref{sec:attention_aggregation} and extraction of the Markov-maps including the flood fill approach in \cref{sec:markov_maps}.
Finally, we introduce the full M2N2 algorithm \cref{sec:markov_as_distance_function}.

%-------------------------------------------------------------------------
%\subsection{\st{Thresholded}{\cblue Truncated} Nearest Neighbor}
\subsection{Truncated Nearest Neighbor} \label{sec:truncated_nn}
In point prompt based segmentation we are given an image $I \in \mathbb{R}^{H \times W \times 3}$ of width $W$ and height $H$ and a set of labeled prompt points $D=\{(x_1, y_1), (x_2, y_2), ..., (x_N, y_N)\}$ where $x_i \in \mathbb{R}^2$ are the 2D spatial coordinates of each prompt point in image pixel space and the labels $y_i\in \{0, 1\}$ denote whether a point $x_i$ belongs to the background $y_i=0$ or foreground $y_i=1$.
To perform k-NN segmentation with $k=1$, we assign each query pixel $x_q$ of our output segmentation the class of its nearest neighbor $x_{i^*}$:
\begin{equation}
    i^* = \underset{i}{\arg \min}~ d(x_i, x_q) ,\;\;\;\;\;\;\;\; y_q = y_{i^*} \label{eqn:basic_nn}
\end{equation}
where $d(\cdot, \cdot): \mathbb{R}^2 \times \mathbb{R}^2 \rightarrow \mathbb{R}_{\geq 0}$ is a semantic distance function measuring the semantic similarity between the query pixel $x_q$ and prompt point $x_i$.
In our case, we allow $d(\cdot, \cdot)$ to be asymmetric, meaning that $d(x_i, x_q)$ may, but does not have to, equal $d(x_q, x_i)$.
Using a canonical 1-NN, a single foreground prompt point would always segment the entire image as foreground, requiring a minimum of two prompts, one foreground and one background prompt, to get a useful segmentation.
To mitigate this limitation and reduce NoC, we further extend the 1-NN algorithm with a distance threshold of $1$ and define $\hat{y}_q$ as:
\begin{equation}
    \hat{y}_q = \begin{cases}
        y_{i^*} & \text{if }  d(x_{i^*}, x_q) \le 1\\
        0 & \text{otherwise.}
    \end{cases}  \label{eqn:truncated_nn}
\end{equation}
If the distance between a query pixel $x_q$ and its nearest neighbor $x_{i^*}$ exceeds the threshold of $1$, it is classified as background $\hat{y}_q=0$, independent of its nearest neighbor's class $y_{i^*}$.
The main challenge is to find a good distance function $d(\cdot, \cdot)$ which measures the semantic dissimilarity of a prompt point $x_i$ and a query pixel $x_q$.

% $d(\cdot, \cdot)$ which should have the following properties:
% \begin{itemize}
%    \item $d(x_i, x_{q_1}) \approx d(x_i, x_{q_2})$ if the points at $x_{q_1}$ and $x_{q_2}$ are in a semantically similar region
%    \item $d(x_i, x_{q_1}) \ll d(x_i, x_{q_2})$ if $x_i$ and $x_{q_1}$ are semantically closer than $x_{i}$ and $x_{q_2}$
%    \item All $x_q$ satisfying $d(x_i, x_q) \le 1$ should form a semantically consistent region.
%\end{itemize}

%-------------------------------------------------------------------------
\subsection{Attention Aggregation} \label{sec:attention_aggregation}
In this paper, we use the pre-trained SD 2.
Given an image $I$, we perform a single denoising step by computing a forward pass through the denoising U-Net and extract the multi-head self-attentions $\boldsymbol{S}_j$ of each transformer block $j$.
Each tensor $\boldsymbol{S}_j$ is of the shape $N_{heads} \times h \times w \times h \times w$.
$N_{heads}$ denotes the number of attention heads and $h$ and $w$ are the height and width of the attention maps respectively.
Each attention map $\boldsymbol{S}_j[n, k, l, :, :]$ of the tensor $\boldsymbol{S}_j$ is a probability distribution, meaning the sum of each map's elements is equal to $1$.
We define the aggregated tensor $\boldsymbol{A} \in \mathbb{R}^{h \times w \times h \times w}$ as follows\footnote{
We do not require any resizing or upscaling since we only consider the $5$ attention tensors illustrated in \cref{fig:framework_overview}. These tensors are of the highest resolution in the U-Net, providing attention tensors $\boldsymbol{S}_j$ of the same shape. We simply average along the attention-head dimension with equal weights since we assume all heads represent very similar attentions~\cite{DiffuseAttendSegment}.}:
\begin{align}
    \boldsymbol{A}[:, :, :, :] &= \sum_{j \in \mathrm{AttnBlocks}} w_{j} \sum_{n}^{N_{heads}} \boldsymbol{S}_j[n, :, :, :, :]
\end{align}
Ensuring that the sum of weights $w_j \in [0, 1]$ is $1$, the result is a single attention tensor $\boldsymbol{A}$ of which each 2D-map $\boldsymbol{A}[k, l, :, :]$ is a probability distribution.

%-------------------------------------------------------------------------
\subsection{From Attention Tensor to Markov-Maps} \label{sec:markov_maps}
First, we flatten the attention tensor $\boldsymbol{A} \in \mathbb{R}^{h \times w \times h \times w}$ to obtain a matrix $A \in \mathbb{R}^{(h \cdot w) \times (h \cdot w)}$.
Since $A$ is a right stochastic matrix\footnote{This is the case because each attention map $\boldsymbol{A}[i, j, :, :]$ is a probability distribution and so is $A[k, :]$}, we can use it as a transition matrix in a Markov chain:
\begin{equation}
    p_t = p_0 \cdot A^t \;\;\;\; \mathrm{with} \;\;\;\; p_0 = onehot(x)
\end{equation}
where the probability distribution row vector $p_0 \in \mathbb{R}^{1 \times (h \cdot w)}$ is the start state and $p_t$ $ \in \mathbb{R}^{1 \times (h \cdot w)}$ is the state after $t \in \mathbb{Z}_{\ge 0}$ iterations.
The start state $p_0$ is the one-hot encoded vector of a prompt point $x$.
$A$ is strictly positive and generally irreducible, aperiodic stochastic matrix.
Therefore, for $t \rightarrow \infty$, each $p_0$ converges to a stationary distribution $p_\infty$ of the Markov chain~\cite{Ross97}.
The state $p_\infty$ depends on the attention matrix $A$ and is therefore different for each image.
In order to be image agnostic, we remove the per-image bias in $p_\infty$, by applying iterative proportional fitting (IPF) to the matrix $A$.
This converts $A$ to a doubly stochastic matrix.
For all images, $A$ has therefore the property that each start state $p_0$ converges to a uniform distribution $p_\infty = \frac{1}{h \cdot w} \boldsymbol{1}$ for $t \rightarrow \infty$~\cite{sinkhornDoublyStochastic}.
As a result, the Markov chain describes the process from the lowest possible entropy in $p_0$ to the highest possible entropy in $p_\infty$\footnote{The one-hot encoded start state has one element set to $1$ and all others set to $0$ and therefore an entropy of $0$. The uniform distribution in $p_\infty$ yields the maximum uncertainty and therefore the largest entropy.}.
Importantly, we are able to control the rate of convergence by changing the entropy of the attention maps $A[k, :]$ by modifying the temperature $T \in \mathbb{R}_{>0}$:
\begin{align}
    A_{new}[k, l] = \frac{\exp\left(\frac{1}{T} \log(A_{prev}[k, l])\right)}{\sum_j \exp\left(\frac{1}{T} \log(A_{prev}[k, j])\right)} \label{eqn:change_temperature}
\end{align}
\cref{eqn:change_temperature} is applied before the IPF to ensure that the transition matrix $A$ is doubly stochastic. Temperatures $T$ higher than $1$ increase the overall entropy of $A$ and therefore require fewer time steps $t$ to reach a uniform distribution in $p_t$.
The key idea is to measure the time $t$ it takes for each element $k$ in $p_0$ to converge to the uniform distribution:
\begin{align}
    m[k] &= \min \left\{t\in \mathbb{Z}_{\ge 0}\ | \ \frac{p_t[k]}{\max p_t} > \tau \right\} \label{eqn:markov_map}
\end{align}
Here $\max p_t$ returns the maximum element of the vector $p_t$ and the vector $m \in \mathbb{Z}_{\ge 0}^{1 \times (h \cdot w)}$ stores the minimal time required of each element $k$ to saturate, i.e., to reach the relative probability threshold $\tau$.
As a result, the hyperparameter $\tau$ allows us to control how quickly the element $k$ saturates.
Since $m$ only contains integers, we also perform a linear interpolation between consecutive time frames to obtain a smoother result.
We then reshape $m$ into a matrix of the shape $h \times w$ and upscale it with JBU~\cite{JBU} to the input image resolution of $H \times W$, resulting in $M \in \mathbb{R}^{H \times W}$.
Since $M$ measures the number of iterations of the Markov chain, we call the resulting map the Markov-iteration-map or in short Markov-map.
\\
\\
\textbf{Flood Fill}.
The task of point prompt based segmentation typically requires instance segmentation.
This proves to be challenging for the proposed Markov-maps, since self-attentions do not take object instances into account.
Therefore, we propose a modified flood fill approach.
Given a prompt point $x$ and its corresponding Markov-map $M$, we perform a flood fill on $M$, using the pixel at $x$ as the starting pixel.
In contrast to classical flood fill, we do not set each pixel to a desired color.
Instead, we store the minimum required flood threshold to reach each pixel.
This simple yet effective approach suppresses local minima and ensures a global minimum at the prompt point $x$.
Our approach requires that the instances do not overlap\footnote{Please note that M2N2 is still able to separate overlapping instances by utilizing multiple prompts.}.
After applying our modified flood fill we obtain the final Markov-map.

\begin{table}[t!]
    \centering
    \begin{tabular}{@{}c@{\hskip 5pt}c@{\hskip 5pt}c@{\hskip 5pt}c@{\hskip 5pt}c@{\hskip 5pt}|@{\hskip 3pt}c@{\hskip 3pt}c@{}}
        \toprule
        
        \multicolumn{5}{c|@{\hskip 3pt}}{Attention Aggregation Weights} & \multicolumn{2}{c}{DAVIS}\\
        $w_{\mathrm{down}0}$ & $w_{\mathrm{down}1}$ & $w_{\mathrm{up}0}$ & $w_{\mathrm{up}1}$ & $w_{\mathrm{up}2}$ & NoC85$\downarrow$ & NoC90$\downarrow$\\
        
        \midrule

        $1$ & $\color{lightgray}{0}$ & $\color{lightgray}{0}$ & $\color{lightgray}{0}$ & $\color{lightgray}{0}$ & $11.34$ & $15.25$\\
        $\color{lightgray}{0}$ & $1$ & $\color{lightgray}{0}$ & $\color{lightgray}{0}$ & $\color{lightgray}{0}$ & $9.38$ & $13.18$\\
        $\color{lightgray}{0}$ & $\color{lightgray}{0}$ & $1$ & $\color{lightgray}{0}$ & $\color{lightgray}{0}$ & $4.86$ & $\underline{6.90}$\\
        $\color{lightgray}{0}$ & $\color{lightgray}{0}$ & $\color{lightgray}{0}$ & $1$ & $\color{lightgray}{0}$ & $\underline{4.79}$ & $7.10$\\
        $\color{lightgray}{0}$ & $\color{lightgray}{0}$ & $\color{lightgray}{0}$ & $\color{lightgray}{0}$ & $1$ & $6.00$ & $9.02$\\

        \midrule
        
        $\color{lightgray}{0}$ & $\color{lightgray}{0}$ & $0.5$ & $0.5$ & $\color{lightgray}{0}$ & $\textbf{4.60}$ & $\textbf{6.72}$\\

        \bottomrule
        
    \end{tabular}
    \caption{\textbf{Ablation study of the attention blocks} on DAVIS. The blocks $\mathrm{up}0$ and $\mathrm{up}1$ achieve the lowest individual NoC.}
    \label{tab:attention_aggregation}
\end{table}

%-------------------------------------------------------------------------
\subsection{M2N2: Markov-Map Nearest Neighbor} \label{sec:markov_as_distance_function}
Using the one-hot encoded coordinates of each prompt point $x_i$ as start states, we generate a \mm $M_i$ for each prompt point $x_i$.
This allows us to construct the following distance function for the truncated nearest neighbor:
\begin{align}
    d(x_i, x_q) = \frac{M_i[x_q]}{\lambda_i} \label{eqn:markov_distance_function}
\end{align}
where $M_i[x_q]$ denotes the value of the Markov-map $M_i$ at the pixel $x_q$.
Since every distance greater than $1$ is assigned as background, we can use the scalar $\lambda_i$ as a threshold to truncate the \mm $M_i$.
Choosing a good threshold $\lambda_i$ is crucial to reduce the NoC. Therefore, we introduce a heuristic to adaptively determine an optimal $\lambda_i$:
\begin{align}
    \lambda_i &= \underset{\lambda}{\arg \max}~ s_i(\lambda)
\end{align}
Given a threshold $\lambda$, we define the total score function $s_i(\lambda)$ as a product of four specific score functions, taking into account a prior, semantic edges and prompt point consistencies:
\begin{align}
    s_i(\lambda) &= 
    s_{i, \mathrm{prior}}(\lambda) \cdot 
    s_{i, \mathrm{edge}}(\lambda) \cdot 
    s_{i, \mathrm{pos}}(\lambda) \cdot 
    s_{i, \mathrm{neg}}(\lambda) \label{eqn:total_score_function_definition}
\end{align}
Each specific score $s_{i, \cdot}(\lambda)$ is based on the individual segmentation  at the threshold $\lambda$ of a single prompt point $x_i$:
\begin{itemize}
    \item $s_{i, \mathrm{prior}}(\lambda)$ = $1$ if the segment size relative to the image size is less than $40\%$, $0$ otherwise
    \item $s_{i, \mathrm{edge}}(\lambda)$ is the average Sobel gradient of $M_i$ computed along the segment boundary pixels
    \item $s_{i, \mathrm{pos}}(\lambda)$ is the percentage of all prompt points $x_j$ having the same class $y_i=y_j$ inside the segment
    \item $s_{i, \mathrm{neg}}(\lambda)$ = $0$ if any other prompt point $x_j$ of wrong class $y_i \ne y_j$ is inside the segment, $1$ otherwise.
\end{itemize}
By inserting \cref{eqn:markov_distance_function} into \cref{eqn:basic_nn} and \cref{eqn:truncated_nn}, we arrive at the final M2N2 equations:
\begin{align}
    i^* &= \underset{i}{\arg \min}~ \frac{M_i[x_q]}{\underset{\lambda}{\arg \max}~ s_i(\lambda)}\\
    \hat{y}_q &= \begin{cases}
        y_{i^*} & \text{if }  \frac{M_{i^*}[x_q]}{\underset{\lambda}{\arg \max}~ s_{i^*}(\lambda)} \le 1\\
        0 & \text{otherwise}
    \end{cases}
\end{align}

%% file: sec/4_experiments.tex
\begin{table}[t!]
    \centering
    \begin{tabular}{@{}c@{\hskip 5pt}c@{\hskip 5pt}c@{\hskip 5pt}c@{\hskip 5pt}|@{\hskip 3pt}c@{\hskip 3pt}c@{}}
        \toprule
        
        \multicolumn{4}{c|@{\hskip 3pt}}{Score Functions $s_{i, \cdot}(\lambda)$} & \multicolumn{2}{c}{DAVIS}\\
        $s_{i,\mathrm{prior}}$ & $s_{i,\mathrm{edge}}$ & $s_{i,\mathrm{pos}}$ & $s_{i,\mathrm{neg}}$ & NoC85$\downarrow$ & NoC90$\downarrow$\\
        
        \midrule

        \xmark & \xmark & \xmark & \xmark & $6.47$ & $10.29$\\
        \xmark & \cmark & \cmark & \cmark & $\underline{4.67}$ & $\underline{6.81}$\\
        \cmark & \xmark & \cmark & \cmark & $7.20$ & $9.99$\\
        \cmark & \cmark & \xmark & \cmark & $4.92$ & $6.87$\\
        \cmark & \cmark & \cmark & \xmark & $5.46$ & $8.66$\\

        \midrule
        
        \cmark & \cmark & \cmark & \cmark & $\textbf{4.60}$ & $\textbf{6.72}$\\

        \bottomrule
        
    \end{tabular}
    \caption{\textbf{Ablation study of threshold score functions} $s_{i, \cdot}(\lambda)$ on DAVIS. The first row utilizes no score with $\lambda_i=0.5$.}
    \label{tab:score_function}
\end{table}

%-------------------------------------------------------------------------
\section{Experiments}\label{sec:experiments}

\definecolor{main_comparison_table_color}{rgb}{0.9, 0.9, 0.9}

\begin{table*}[h!]
  \centering
  %\begin{tabular}{@{}ll|@{\hskip 3pt}c@{\hskip 3pt}c@{\hskip 3pt}|@{\hskip 3pt}c@{\hskip 3pt}c@{\hskip 3pt}|@{\hskip 3pt}c@{\hskip 3pt}c@{\hskip 3pt}|@{\hskip 3pt}c@{\hskip 3pt}c@{}}
  \begin{tabular}{>{\kern-\tabcolsep}ll@{\hskip 5pt}c@{\hskip 3pt}c@{\hskip 12pt}c@{\hskip 3pt}c@{\hskip 12pt}c@{\hskip 3pt}c@{\hskip 12pt}c@{\hskip 3pt}c<{\kern-\tabcolsep}}
  
    \toprule
    
%    \multirow{2}{*}{\textbf{Method}} & \multirow{2}{*}{\textbf{Backbone}} & \multicolumn{2}{c|@{\hskip 3pt}}{GrabCut~\cite{grabcut}} & \multicolumn{2}{c|@{\hskip 3pt}}{Berkeley~\cite{berkeleyDataset}} & \multicolumn{2}{c|@{\hskip 3pt}}{SBD~\cite{SBD-Dataset}} & \multicolumn{2}{c}{DAVIS\cite{DAVIS_DAtaset}} \\
    \multirow{2}{*}{Method} & \multirow{2}{*}{Backbone} & \multicolumn{2}{@{}c@{\hskip 15pt}}{GrabCut~\cite{grabcut}} & \multicolumn{2}{@{}c@{\hskip 15pt}}{Berkeley~\cite{berkeleyDataset}} & \multicolumn{2}{@{}c@{\hskip 15pt}}{SBD~\cite{SBD-Dataset}} & \multicolumn{2}{c}{DAVIS~\cite{DAVIS_DAtaset}} \\
    & & NoC85$\downarrow$ & NoC90$\downarrow$ & NoC85$\downarrow$ & NoC90$\downarrow$ & NoC85$\downarrow$ & NoC90$\downarrow$ & NoC85$\downarrow$ & NoC90$\downarrow$\\
    
    \midrule
    \rowcolor{main_comparison_table_color}
    \multicolumn{10}{l}{\textbf{Supervised}: Trained on Ground Truth Labels}\\
    SimpleClick \cite{SimpleClick} & ViT-B & $1.40$ & $1.54$ & $1.44$ & $2.46$ & $3.28$ & $5.24$ & \underline{$4.10$} & $5.48$ \\
    SimpleClick                    & ViT-H & $\textbf{1.32}$ & $1.44$ & $\textbf{1.36}$ & \underline{$2.09$} & $\textbf{2.51}$ & $\textbf{4.15}$ & $4.20$ & $5.34$ \\ 
    CPlot~\cite{clickPromptOptimalTransport} & ViT-B & \underline{$1.34$} & $1.48$ & \underline{$1.40$} & $2.18$ &  \underline{$3.05$} & $4.95$ & $\textbf{4.00}$ & $5.29$ \\
    CFR-ICL~\cite{CFR-ICL} & ViT-H & - & \underline{$1.42$} & - & $\textbf{1.74}$ & - &  \underline{$4.45$} & - & \underline{$4.77$} \\   
    FocalClick~\cite{FocalClick}  &  SegFormerB0-S2 & - & $1.90$ & - & $3.14$ & $4.34$ & $6.51$ & - & $7.06$\\ 
InterFormer-Tiny~\cite{interformer} & ViT-L & - & \textbf{1.36} & - & $2.53$ & $3.25$ & $5.51$ & - & $5.21$\\%0.50(512) & 
OIS~\cite{OISegmentation} & ViT-B & - & - & - & - & - & - & - & \textbf{3.80}\\
    \midrule
    \midrule

    \rowcolor{main_comparison_table_color}
    \multicolumn{10}{l}{\textbf{Unsupervised}: Trained on Pseudo Labels}\\
    TokenCut$^\star$~\cite{TokenCutSO} & ViT-B & $3.04$ & $5.74$ & $6.30$ & $9.97$ & $10.82$ & $13.16$ & $10.56$ & $15.01$\\
    FreeMask$^\star$~\cite{FreeMask} & ViT-B & $4.06$ & $6.10$ & $5.55$ & $9.02$ & $8.61$ & $12.01$ & $8.26$ & $13.05$\\
    DSM$^\star$~\cite{DSM} & ViT-B & $3.64$ & $4.64$ & $5.49$ & $7.75$ & $8.59$ & $11.57$ & $7.08$ & $10.11$\\
    MIS~\cite{MIS} & ViT-B & $1.94$ & $2.32$ & $3.09$ & $\underline{4.58}$ & $\textbf{6.91}$ & $\textbf{9.51}$ & $6.33$ & $8.44$\\

    \rowcolor{main_comparison_table_color}
    \multicolumn{10}{l}{\textbf{Unsupervised}: Training-Free}\\
    GraphCut~\cite{grabcut} & N/A & $7.98$ & $10.00$ & - & $14.22$ & $13.60$ & $15.96$ & $15.13$ & $17.41$\\
    Random Walk~\cite{randomWalkSegmentation} & N/A & $11.36$ & $13.77$ & - & $14.02$ & $12.22$ & $15.04$ & $16.71$ & $18.31$\\
    Geodesic Matting~\cite{geodesicMattingSegmentation} & N/A & $13.32$ & $14.57$ & - & $15.96$ & $15.36$ & $17.60$ & $18.59$ & $19.50$\\
    GSC~\cite{geodesicStartSegmentation} & N/A & $7.10$ & $9.1$2 & - & $12.57$ & $12.69$ & $15.31$ & $15.35$ & $17.52$\\
    ESC~\cite{geodesicStartSegmentation} & N/A & $7.24$ & $9.20$ & - & $12.11$ & $12.21$ & $14.86$ & $15.41$ & $17.70$\\ %\hdashline

    %\midrule

    \rowcolor{highlight}
    Attention-NN & SD2 & $5.48$ & $7.16$ & $6.45$ & $9.61$ & $11.67$ & $14.97$ & $9.66$ & $13.12$\\
    \rowcolor{highlight}
    KL-NN & SD2 & $3.80$ & $5.32$ & $5.11$ & $7.71$ & $10.78$ & $14.00$ & $7.60$ & $10.89$\\
    \rowcolor{highlight}
    M2N2 w/o flood fill & SD2 & $3.06$ & $4.50$ & $4.80$ & $6.51$ & $9.16$ & $12.26$ & $5.50$ & $\underline{7.42}$\\
    \rowcolor{highlight}
    M2N2 & ViT-B & $1.92$ & $3.40$ & $3.55$ & $5.68$ & $7.94$ & $11.53$ & $6.85$ & $10.86$\\
    \rowcolor{highlight}
    M2N2 & SD1.1 & $\underline{1.84}$ & $\underline{2.20}$ & $\underline{2.90}$ & $4.93$ & $8.56$ & $11.66$ & $\underline{5.16}$ & $7.80$\\
    \rowcolor{highlight}
    M2N2 (Ours) & SD2 & $\textbf{1.62}$ & $\textbf{1.90}$ & $\textbf{2.45}$ & $\textbf{3.88}$ & $\underline{7.72}$ & $\underline{10.94}$ & $\textbf{4.60}$ & $\textbf{6.72}$\\
    
    \bottomrule
  \end{tabular}
  \caption{\textbf{Main comparison} of M2N2 with baselines and previous work on all four datasets. $\star$ indicates pseudo label generation methods used on SBD images for training an unsupervised SimpleClick model~\cite{MIS}. All supervised methods were trained on SBD, except InterFormer (COCO~\cite{COCO}+LVIS~\cite{LVIS}) and OIS (HQSeg44K~\cite{HQSAM}). Among the unsupervised methods, M2N2 obtains the best results in three out of four datasets, despite being training-free.}
  \label{tab:main_comparison}
\end{table*}

\textbf{Datasets.} We evaluate M2N2 on 4 public datasets:
\begin{itemize}
    \item \textbf{GrabCut}~\cite{grabcut}: 50 images; 50 instances.
    \item \textbf{Berkeley}~\cite{berkeleyDataset}: 96 images; 100 instances. Some of the images are shared with the GrabCut dataset.
    \item \textbf{SBD}~\cite{SBD-Dataset}: 2857 images; 6671 instances (validation set).
    \item \textbf{DAVIS}~\cite{DAVIS_DAtaset}: 345 images; 345 instances. The segmentation is of very high-quality and instances include many objects with fine details and small structures. We use the same 345 images as used in \cite{SimpleClick, MIS}.
\end{itemize}
\textbf{Evaluation Metrics}. Following previous work \cite{SimpleClick, MIS}, we evaluate our approach by simulating user interaction, in which we place the next click in the center of the largest error region.
The maximum number of clicks for each instance is $20$ and we provide two metrics, the average number of clicks required to reach an IoU of $85\%$ as NoC85 and an IoU of $90\%$ as NoC90.
\\
\\
\textbf{Comparison with Previous Work}.
\cref{tab:main_comparison} compares our approach M2N2 with previous supervised and unsupervised approaches.
To the best of our knowledge, M2N2 is the first unsupervised interactive point prompt based segmentation framework utilizing a pre-trained model without requiring any additional training.
All other methods either are not based on any deep learning, e.g., GrabCut and related ones, or require the generation of pseudo-labels to train an interactive model, e.g., MIS.
Our method surpasses the previous state-of-the-art unsupervised method MIS, which is trained on pseudo labels, on both metrics in three out of four test datasets.
We observe the largest improvement on the DAVIS dataset, where we reduced the NoC85 by $1.73$ and the NoC90 by $1.72$ clicks.
We achieve second best results in SBD.
A possible explanation for this is that all deep-learning models listed in \cref{tab:main_comparison} are trained on the training set of SBD and therefore might have an advantage on this dataset.
\\
\\
\textbf{Baselines}.
We provide two additional baselines in \cref{tab:main_comparison} which use the same framework as M2N2 but without Markov-maps.
Attention Truncated Nearest Neighbor (Attention-NN) uses attention maps as a semantic distance measure.
KL-Divergence Truncated Nearest Neighbor (KL-NN) utilizes a symmetric KL-Divergence between the attention map of the prompt point and all attention maps in the attention tensor as distance function.
Finally, we also provide a version of M2N2 without flood fill.
We observe that the combination of \mms with Flood fill achieves the best results with SD2 as our backbone compared to SD1.1 \cite{StableDiffusion} and ViT-B \cite{DINOv2}.
\\
\\
\textbf{SD2 Domain Bias}.
We notice a significant domain bias of M2N2 on medical data.
Evaluating the IoU's for 10 clicks on BraTS~\cite{BraTS} (SimpleClick: $87\%$, M2N2: $70\%$) and OAIZIB~\cite{OAIZIB} (SimpleClick: $76\%$, M2N2: $66\%$) shows significant bias towards natural images.

%-------------------------------------------------------------------------
\subsection{Implementation Details}
We use the SD2 implementation and weights provided by the Hugging Face transformers package.
We don't add noise to the encoded image latent to keep the results deterministic and perform the single denoising step with empty text prompts.
Due to the large memory requirements of the attention tensors, we run SD on 16-bit floating-point precision and convert it to 32-bit floating-point for the attention aggregation and further processes in our framework.
We implement JBU as described in \cite{JBU} with two modifications.
We change the low-resolution solution sampling from sparse to dense sampling for smoother results and extend the range term to an isotropic Gaussian to better utilize RGB information in the images.
We set $\sigma_\mathrm{spatial}=1$ and $\sigma_\mathrm{range}=0.1$ for RGB color values in the range $[0, 1]$.
For M2N2 we choose the attention tensors of the size $128 \times 128 \times 128 \times 128$ and the SD time step of $100$.
We use the temperature $T=0.65$ together with a relative probability threshold $\tau=0.3$ to compute the \mms with a maximum of $1000$ iterations.
By caching the attention tensor and \mms of previous clicks, we observe an average of $0.6$ seconds per click (SPC) for an image resolution of $854 \times 480$ on a RTX 4090, enabling a near real-time interactive process with the user. For more information on the other backbones please see \cref{sec:additional_backbones_in_detail}.

%-------------------------------------------------------------------------
\subsection{Ablation Study}
\begin{figure*}
  \centering{\includegraphics[width=\textwidth]{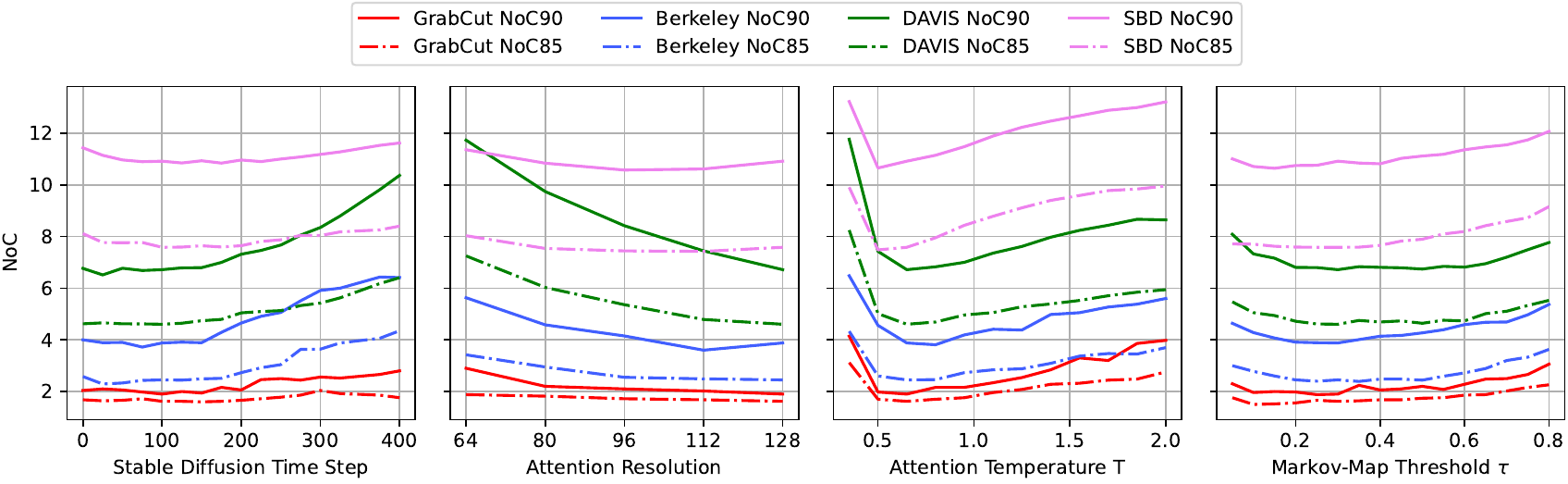}}
  \caption{\textbf{Impact of the hyperparameters of SD and Markov-map}, respectively, on all four datasets, each represented by a single color. Dashed lines correspond to NoC85, continues lines to NoC90. The graph of SBD is based on a randomly sampled subset of 500 images.}
  \label{fig:ablation_four_parameters_on_all_datasets}
\end{figure*}

We perform extensive ablation studies on the hyperparameters of our segmentation algorithm to demonstrate the impact of each component of M2N2.
\\
\\
\textbf{Attention Aggregation}. 
Our experiments in \cref{fig:ablation_four_parameters_on_all_datasets} show that higher attention tensor resolution improves NoC significantly for most of our datasets, performing best at a resolution of $128 \times 128 \times 128 \times 128$.
This is surprising since it requires an input image size of $1024 \times 1024$ which is beyond SD2's training resolution of $768 \times 768$\footnote{Due to memory requirements we did not test higher resolutions.}.
We also evaluate the NoC for the SD time step, which is required for the single denoising step.
Time steps greater than $200$ increase the NoC which we assume is due to the distribution shift caused by not adding noise to the encoded latent.
For the aggregation of attention tensors, we evaluate each attention block individually in \cref{tab:attention_aggregation} and observe that the attention tensors of up0 and up1 individually achieve significantly better NoC than the other layers.
Aggregating up0 and up1 results in the best NoC.
\\
\\
\textbf{Markov-maps}.
Lowering the temperature $T$ of the aggregated tensor $A$ gradually improves the NoC in \cref{fig:ablation_four_parameters_on_all_datasets} up to $T=0.5$. Smaller values of $T<0.5$ reduce the entropy and therefore require more iterations in the Markov chain, exceeding the maximum number of iterations and causing numerical instabilities.
Different settings of the relative probability threshold $\tau$ prove to have a relatively low impact on the NoC.
\\
\\
\textbf{Truncated Nearest Neighbor}. Experiments in \cref{tab:score_function} show the contribution of each score function $s_{i, \cdot}(\lambda)$ to the NoC of the total score function $s_i(\lambda)$.
In the second row we observe that the function $s_{i, \mathrm{prior}}$, has the lowest impact on the NoC.
Removing $s_{i, \mathrm{edge}}$ on the other hand increases NoC90 to $9.99$ and is therefore the most important score function.

\begin{figure}
  \centering
  \includegraphics[width=\columnwidth]{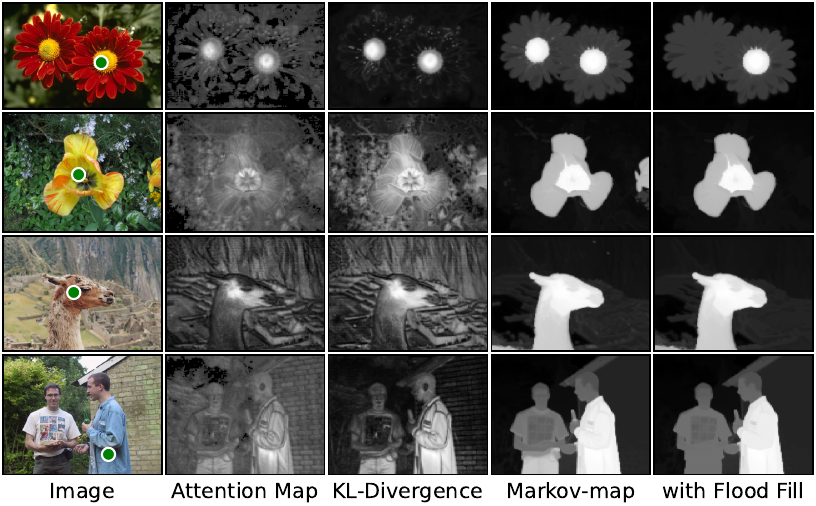}
  \caption{\textbf{Comparison of semantic maps}. Each map is generated from a single prompt point. For better comparison, Markov-maps are inverted such that the lowest value is white and the highest value is black.}
  \label{fig:ablation_attention_vs_markov_map}
\end{figure}

\begin{figure*}
  \centering{\includegraphics[width=\textwidth]{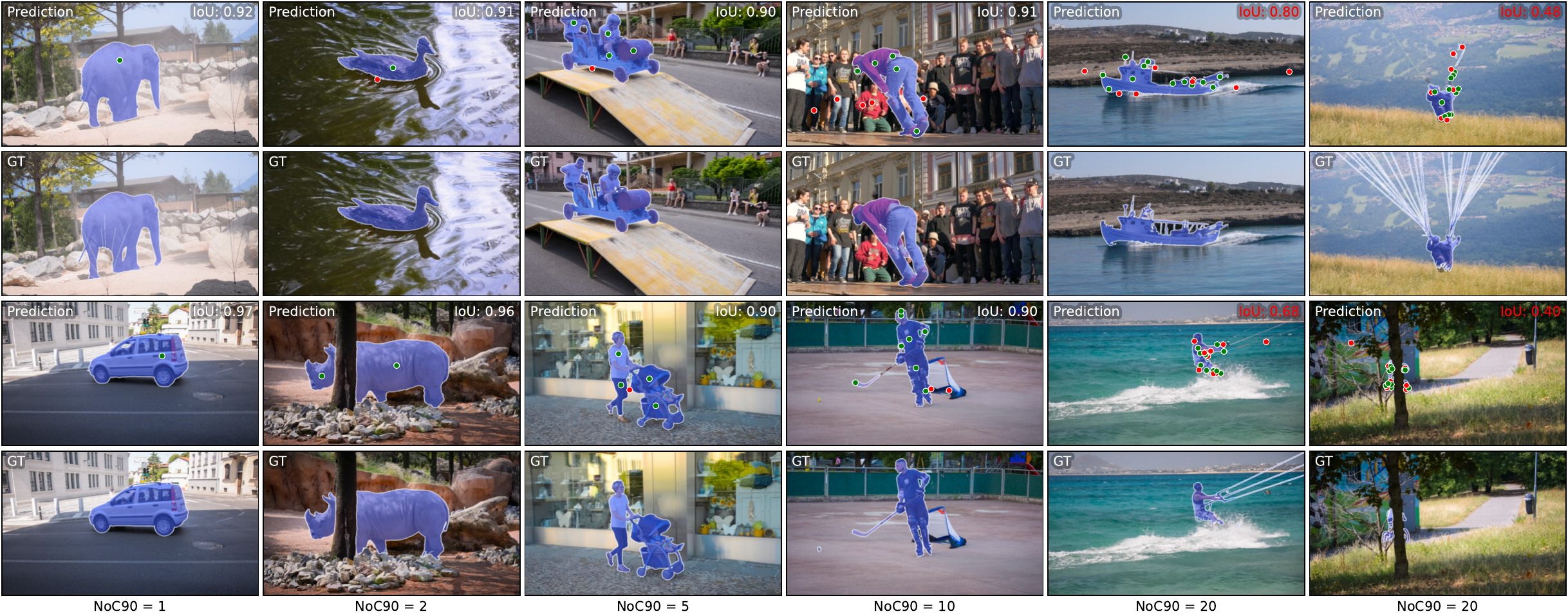}}
  \caption{\textbf{Segmentation examples on DAVIS \cite{DAVIS_DAtaset}}. Each column shows examples slected on the NoC90 value, ranging from easy cases $\mathrm{NoC90}=1$ on the left to difficult cases $\mathrm{NoC90}=10$ and failure cases $\mathrm{NoC90}=20$ on the right. Foreground points are shown in green and background points in red. The bottom right example is especially difficult for M2N2 by only having small thin isolated structures.}
  \label{fig:ablation_davis_examples}
\end{figure*}

%-------------------------------------------------------------------------
\subsection{Qualitative Results and Limitations of M2N2} \label{sec:qualitative_results_of_m2n2}
\cref{fig:ablation_attention_vs_markov_map} compares our proposed Markov-maps with other semantic maps obtained from a single prompt point.
We observe that Markov-maps are less noisy, have clearer semantic boundaries, and have more uniform values within semantically similar regions.
Additionally, we notice that Markov-maps nicely reflect a semantic hierarchy due to the segment size ambiguity of a single prompt point.
For example, in the first row and right-most column, the two overlapping flowers have a hierarchy of three levels.
The first level contains the right flower's ovary, the second contains both flowers together and the final level covers everything.
This example also shows the strengths and weaknesses of our flood fill approach.
As the ovaries are not overlapping, it enables instance segmentation of the right flower's ovary by suppressing the local minimum of the left flower's ovary.
On the other hand, flood fill does not separate the right flower individually, because both flowers are semantically equal and overlapping.
Please note that this limitation only applies to Markov-maps of a single point.
M2N2 is still able to segment overlapping instances with the truncated nearest neighbor of multiple prompt points. (See \cref{fig:score_metrics_graphs})

Examples from the DAVIS dataset in \cref{fig:ablation_davis_examples} show a similar issue in the $\mathrm{NoC90}=10$ column, where a single dancer is in front of a semantically similar crowd, requiring multiple corrective background points for proper separation.
Another limitation of flood fill is that obstructions can split instances into multiple areas.
For example, the segmentation of the obstructed rhinoceros requires two points, instead of one.
Due to the limited attention resolution, M2N2 faces challenges in segmenting thin and fine structures, as shown in the 2 rightmost columns.
The pixel-level truncated nearest neighbor still enables the segmentation of such structures but requires significantly more prompt points to reach the desired accuracy.
In general, we observe that M2N2 generates consistent segments with sharp semantic boundaries without being trained on any segmentation labels.

%% file: sec/conclusions.tex
\section{Conclusion}\label{sec:conclusions}
We proposed M2N2, a novel method for unsupervised training-free point prompt based interactive segmentation.
By interpreting an aggregated self-attention tensor of Stable Diffusion 2 as a Markov transition operator, we generated semantically rich \mms.
We showed that \mms have less noise, clearer semantic boundaries, and more uniform values for semantically similar regions.
By combining Markov-maps with truncated nearest neighbor, we developed M2N2, which even outperformed \sota~unsupervised training-based methods in most of the datasets.
Current limitations are the segmentation of fine structures due to low attention resolution, as well as overlapping or obstructed segments, where M2N2 may require more prompt points. 
% Therefore, future work may involve using higher attention resolutions and improving instance separation.%different prompting methods such as text prompts.

%\textbf{Limitations.} 

%\textbf{Conclusion.}

%% file: sec/X_suppl.tex
\clearpage
\setcounter{page}{1}
\maketitlesupplementary

\begin{figure*}
  \centering{\includegraphics[width=\textwidth]{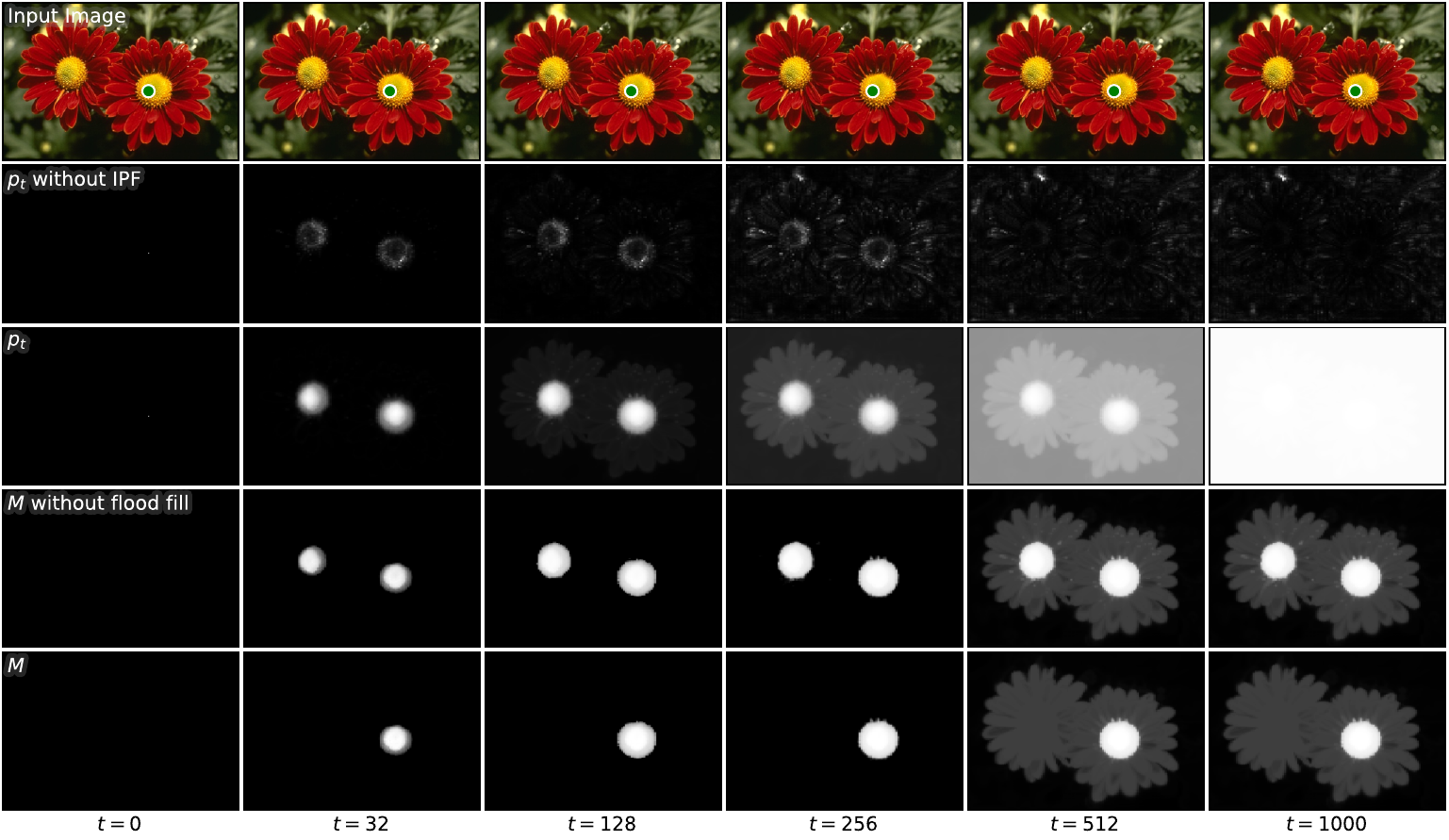}}
  \caption{\textbf{Generation process of a Markov-map}. Each column shows the current state of the probability distribution $p_t$ and the corresponding Markov-map $M$ for a given number of iterations $t$. The first row contains the input image and prompt point. The second and third row show the probability distributions $p_t$ of the original attention tensor without IPF and the doubly stochastic attention tensor resulting from applying IPF. The last two rows are the Markov-maps $M$ with and without using flood fill. The Markov-maps are shown with the maximum number of iterations set to $t$. Each map is scaled up to the image resolution with nearest-neighbor interpolation instead of JBU for better comparison.}
  \label{fig:markov_process_visualization}
\end{figure*}

\begin{figure*}
  \centering{\includegraphics[width=\textwidth]{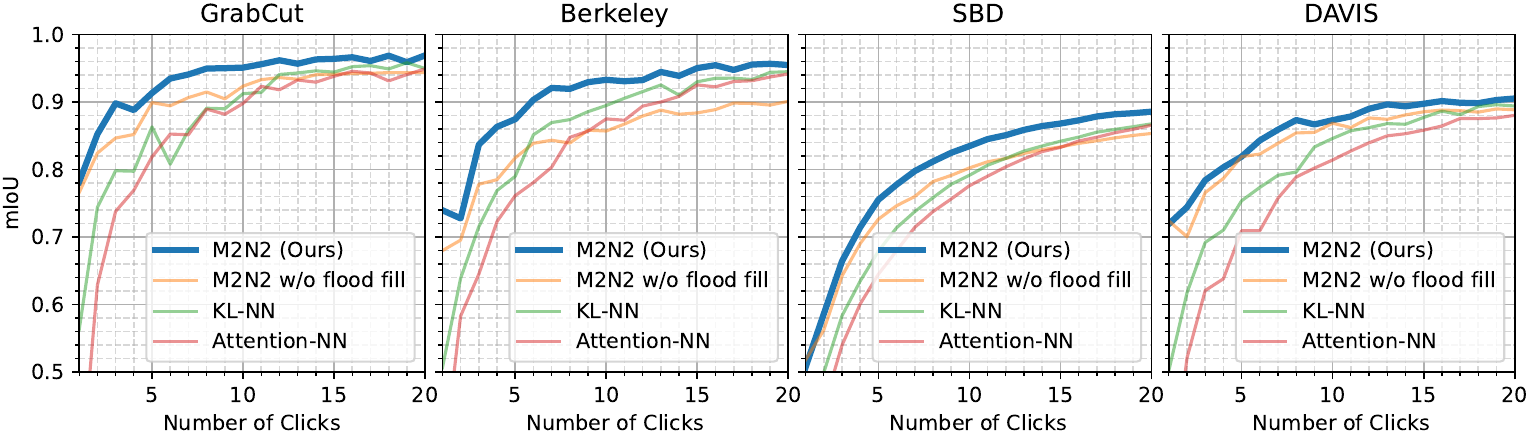}}
  \caption{\textbf{mIoU per NoC}. For each dataset we show the mIoU at a given number of clicks.}
  \label{fig:mIoU_per_NoC}
\end{figure*}

\begin{figure*}
  \centering{\includegraphics[width=\textwidth]{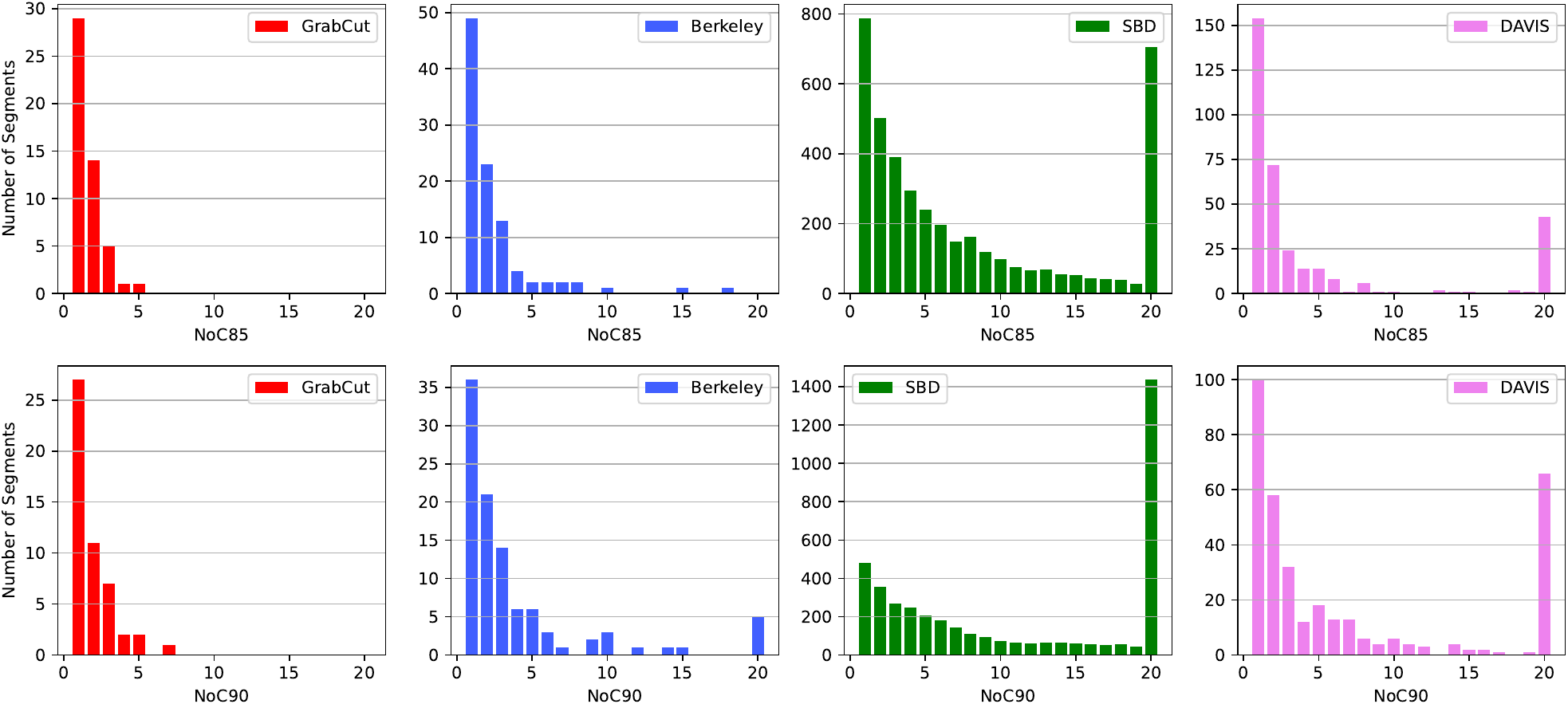}}
  \caption{\textbf{Distribution of the NoC} for each dataset. The maximum number of clicks is set to 20.}
  \label{fig:NoC_distributions}
\end{figure*}

\begin{figure*}
  \centering{\includegraphics[width=\textwidth]{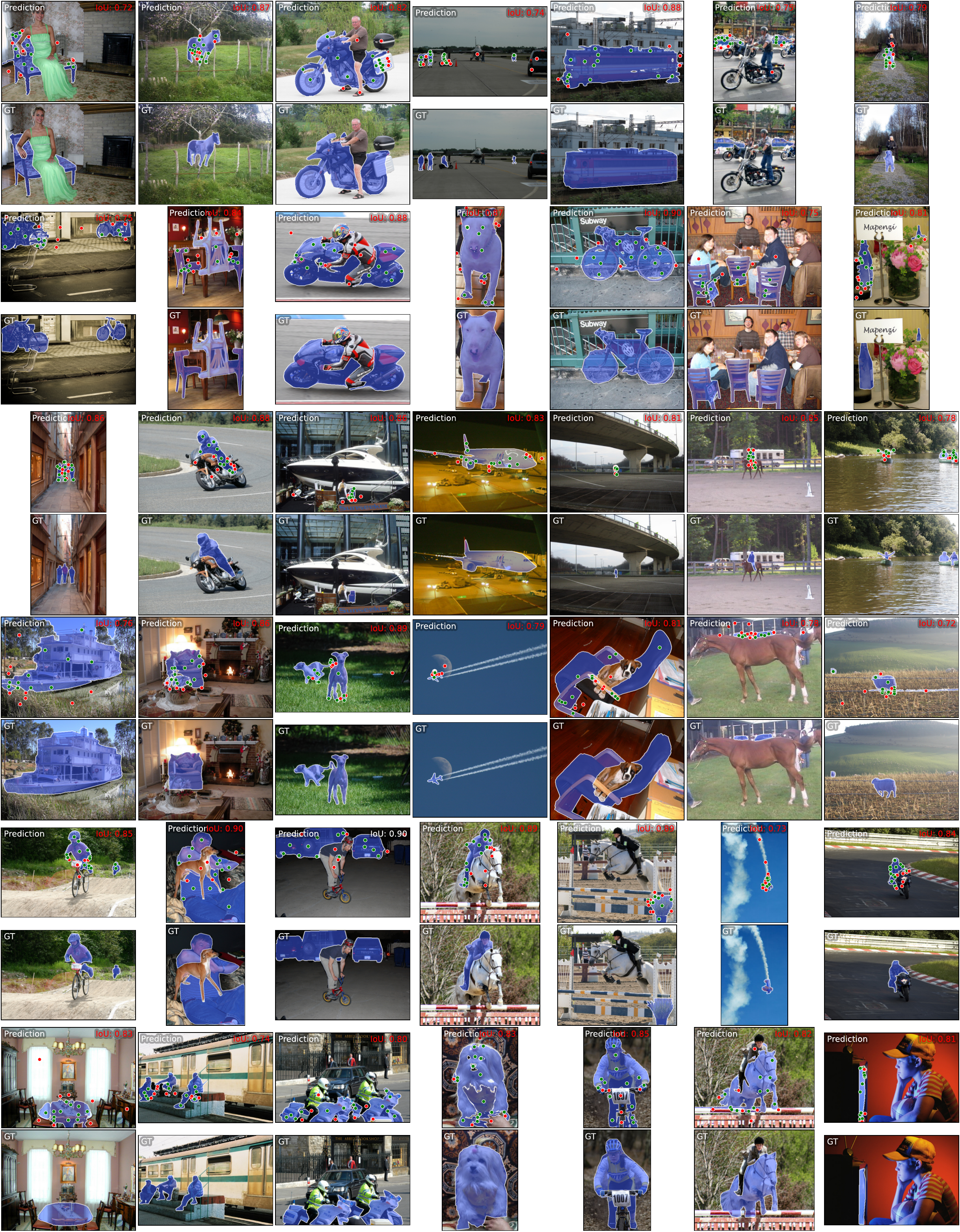}}
  \caption{\textbf{42 randomly sampled failure cases on SBD}. All examples here have $\mathrm{NoC90}=20$.}
  \label{fig:SBD_failure_examples}
\end{figure*}

\begin{figure}
  \centering{\includegraphics[width=\columnwidth]{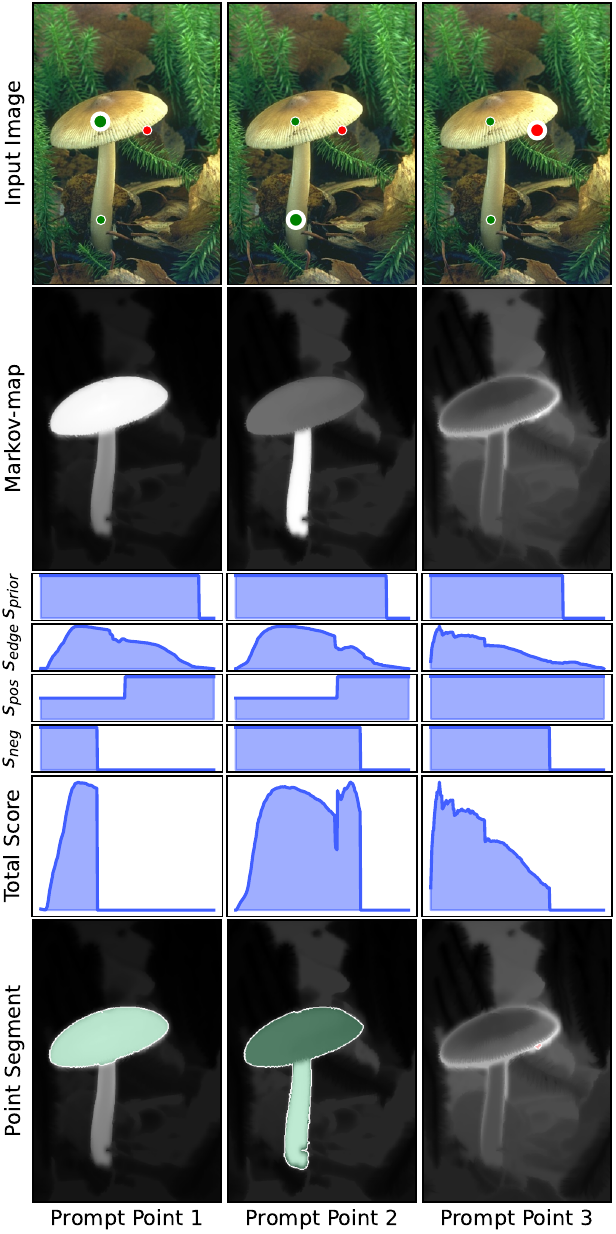}}
  \caption{\textbf{Visualization of individual score functions}. For each prompt point $i$, we show individual score functions $s_{i, \mathrm{prior}}(\lambda)$ and the total score function $s_i(\lambda)$. The x-axis of each graph is the threshold and the respective score is on the y-axis. The last row shows the segmentation resulting from the threshold of the maximum  total score $s_i(\lambda)$.}
  \label{fig:score_metrics_graphs_simple}
\end{figure}

\begin{figure*}
  \centering{\includegraphics[width=\textwidth]{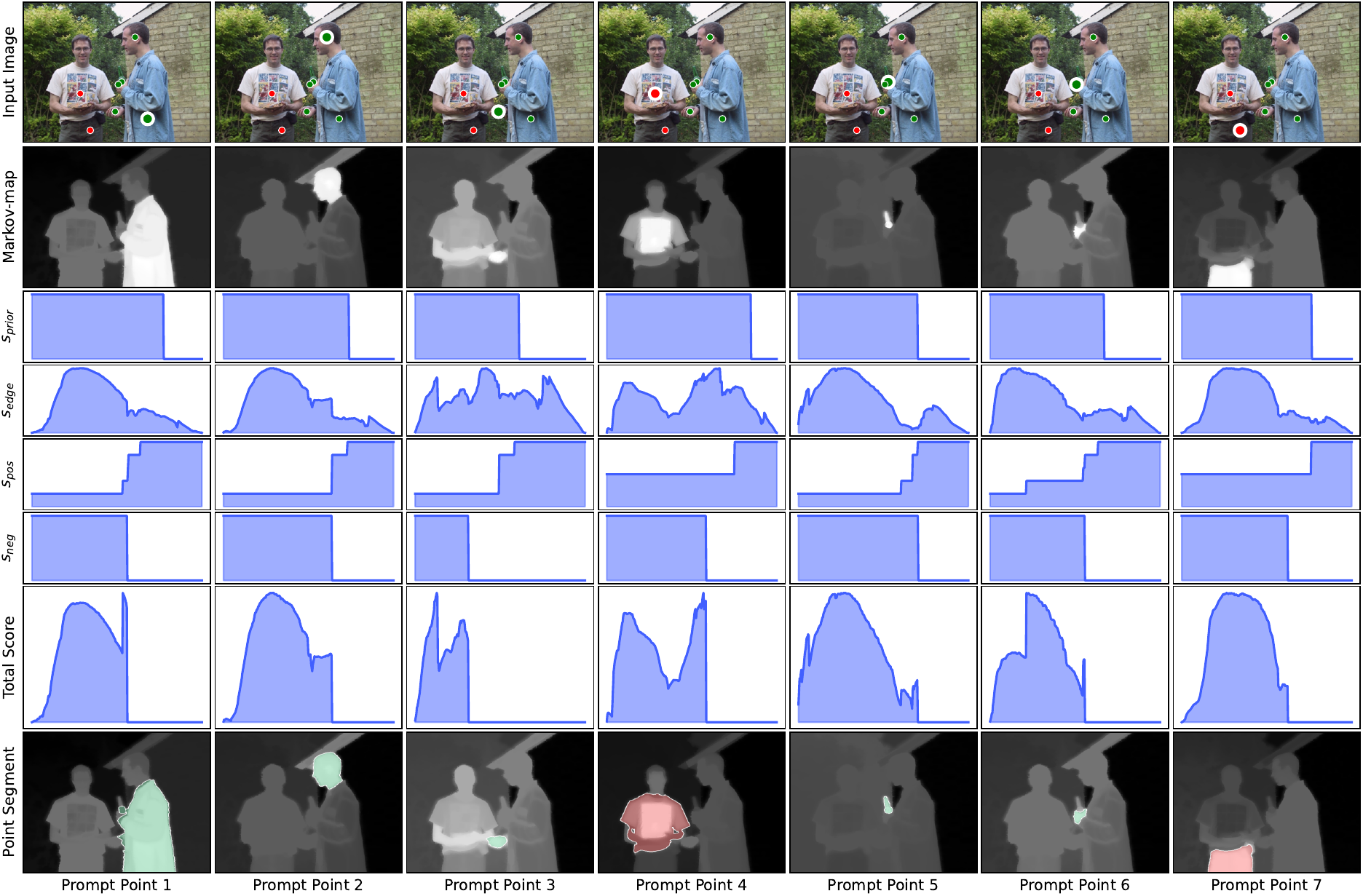}}
  \caption{\textbf{Complex example of score functions}. The example segmentation utilizes 7 prompt points. We note that the each prompt point's segment has a clear semantic meaning (starting from left): jacket, head, hand, shirt, bottle, hand, trousers. It also shows that individual Markov-maps can segment small objects.}
  \label{fig:score_metrics_graphs}
\end{figure*}

\begin{figure*}
  \centering{\includegraphics[width=\textwidth]{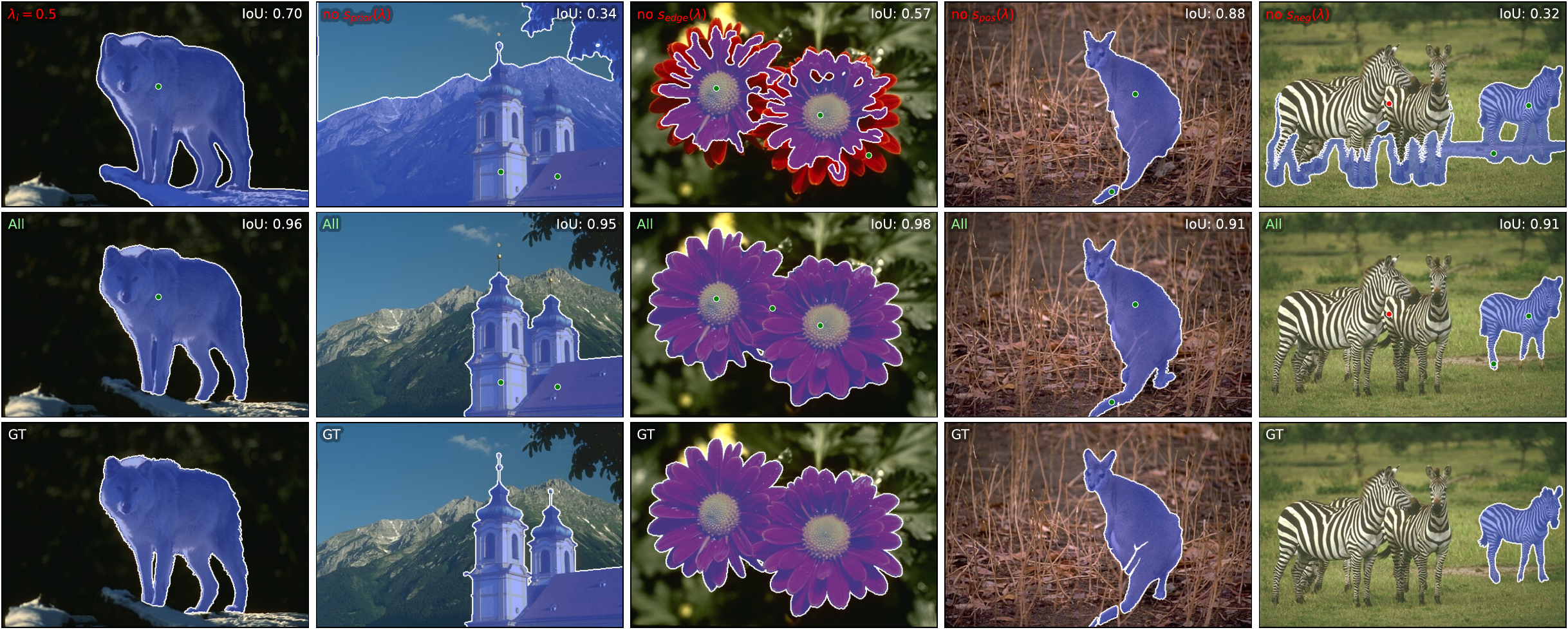}}
  \caption{\textbf{Qualitative examples of score functions} $s_{i, \cdot}(\lambda)$. The first column fixes the threshold to a constant $\lambda_i=0.5$, effectively using no score functions $s_{i, \cdot}(\lambda)$, while the other four columns show the impact of disabling one of the four functions. The second and third row display the segmentation obtained by utilizing all score functions and the ground truth.}
  \label{fig:ablation_figure_scores}
\end{figure*}

\section{Adjusting Temperature after the Softmax Operation}
The softmax of logits $v \in \mathbb{R}^N$ and temperature $T \in \mathbb{R}_{>0}$ is given by the following equation:
\begin{align}
    p_i = \frac{e^{\frac{1}{T} v_i}}{\sum_j e^{\frac{1}{T} v_j}}
\end{align}
where $p \in \mathbb{R}^N$ is the resulting probability distribution. In our approach, the aggregated attention tensor contains the output probabilities of the self-attention softmax operations. Therefore, we take the logarithm of $p$ to obtain logits and then apply the softmax with temperature $T$ to get $p'$. This is mathematically equivalent to applying the temperature $T$ directly on the logits $v$:
\begin{align}
    p'_i &= \frac{e^{\frac{1}{T} \log(p_i)}}{\sum_j e^{\frac{1}{T} \log(p_j)}}\\
    &= \frac{e^{\frac{1}{T} \log(\frac{e^{v_i}}{\sum_j e^{v_j}})}}{\sum_j e^{\frac{1}{T} \log(\frac{e^{v_j}}{\sum_k e^{v_k}})}}\\
    &= \frac{e^{\frac{1}{T} v_i - \log(\sum_j e^{v_j})}}{\sum_j e^{\frac{1}{T} v_j - \log(\sum_k e^{v_k})}}\\
    &= \frac{\frac{e^{\frac{1}{T} v_i}}{\sum_j e^{v_j}}}{\frac{\sum_j e^{\frac{1}{T} v_j}}{\sum_k e^{v_k}}}\\
    &= \frac{e^{\frac{1}{T} v_i}}{\sum_j e^{\frac{1}{T} v_j}} \cdot \frac{\sum_k e^{v_k}}{\sum_j e^{v_j}}\\
    &= \frac{e^{\frac{1}{T} v_i}}{\sum_j e^{\frac{1}{T} v_j}}
\end{align}

\section{Pseudocode for Flood Fill Approach}
As described in \cref{sec:markov_maps}, our modified flood fill approach does not store a desired color in the output map, but instead stores the minimum required flood fill threshold to reach each pixel.
We show a possible implementation of this algorithm in \cref{alg:flood_fill}.
The proposed algorithm is sequential for each image pixel and therefore the runtime scales at least $\ge O(H \cdot W)$ with the image resolution.
We implement the algorithm on the CPU and it has the largest impact on the per per-click processing time.
\renewcommand{\algorithmicrequire}{\textbf{Input:}}  % Redefine to show 'Input:'
\renewcommand{\algorithmicensure}{\textbf{Output:}}   % Redefine to show 'Output:'
\begin{algorithm}
\caption{Modified Flood Fill}
\begin{algorithmic}[1]
\REQUIRE Prompt point pixel $x$ and the corresponding Markov-map $M$
\ENSURE Updated Markov-map $M'$
\STATE Initialize empty $M' \in \mathbb{R}^{H \times W}$
\STATE Initialize priority queue $q \leftarrow \{(0, x)\}$ 
\WHILE{$q$ is not empty}
    \STATE Pop $\lambda'$, $x'$ pair with lowest threshold $\lambda'$ from $q$ 
    \STATE $M'[x'] \gets \lambda'$
    \FOR{horizontal and vertical neighbor pixels $y$ of $x'$}
        \IF{$M'[y]$ is empty and $y$ not in $q$}
            \STATE $\lambda \gets \max(\lambda', \mathrm{abs}(M[y] - M[x]))$
            \STATE $q \gets q \cup \{(\lambda, y)\}$
        \ENDIF
    \ENDFOR
\ENDWHILE
\RETURN $M'$
\end{algorithmic}
\end{algorithm} \label{alg:flood_fill}

\section{Hyperparameters of the Baselines}
In \cref{tab:main_comparison} we compared our M2N2, which utilizes Markov-maps, with the raw attention maps in Attention-NN and the KL-Divergence in KL-NN. We choose different hyperparameters for each of these two baselines to improve their NoC. For Attention-NN we set the temperature $T=10$ and for the KL-NN we set $T=2$ and clipped attention values to a range of $[10^{-5}, 1]$.

\section{Visualization of the Markov chain}
Our Markov-maps introduced in \cref{sec:markov_maps} use the attention tensor $A$ as a Markov transition operator to create a Markov chain.
In \cref{fig:markov_process_visualization} we show an example of the probability distribution $p_t$ and Markov-map $M$ over time.
The first row shows the distribution $p_t$ for the attention tensor without applying IPF.
As expected, $p_t$ does not converge to a uniform distribution in $p_{1000}$ but instead converges to the image-dependent terminal state as mentioned in \cref{sec:markov_maps}.
By applying IPF we convert the attention tensor to a doubly stochastic matrix and therefore obtain an image agnostic uniform distribution in $p_{1000}$.
The last two rows show the resulting Markov-maps for a given maximum number of iterations.
The Markov-map in this example is already fully generated after $377$ iterations with a attention temperature of $T=0.65$ and a relative probability threshold of $\tau = 0.3$.
In general, lowering the temperature $T$ requires significantly higher number of iterations in the IPF to obtain a doubly stochastic matrix, and the Markov chain to converge to the uniform distribution.

\section{Additional Results}
\cref{fig:mIoU_per_NoC} shows the convergence of M2N2 on the GrabCut, Berkeley, SBD and DAVIS datasets, respectively.
On all four datasets, M2N2 converges faster than Attention-NN, KL-NN and M2N2 method without flood fill.
We observe the fastest convergence on the GrabCut dataset and the slowest convergence on the SBD dataset.

In \cref{fig:NoC_distributions}, we show the distribution of NoC for each dataset.
On GrabCut, Berkeley and DAVIS, we observe that M2N2 is able to segment the majority of segments in only a few clicks for both NoC85 and NoC90.
For DAVIS and especially SBD, we find a significantly higher number of failure cases $\mathrm{NoC}=20$.
As discussed in \cref{sec:qualitative_results_of_m2n2}, the high failure rate in the DAVIS datasets is likely due to very thin and fine structures, probably due to the limited self-attention map resolution of $128 \times 128$.
SBD has the highest failure rate and as shown in \cref{tab:main_comparison} the NoC85 and NoC90 on SBD are relatively high in comparison to MIS~\cite{MIS}.
We therefore provide a set of randomly sampled failure cases in \cref{fig:SBD_failure_examples}.
In the following, potential reasons for the relatively low performance of M2N2 on SBD are discussed:
\begin{itemize}
    \item \textbf{High boundary sensitivity:} M2N2 is very sensitive to the semantic boundaries of objects due to the hierarchical structure of the Markov-maps. Since ground truth data in SBD is often represented by polygonal shapes, M2N2 might have difficulties with edges that are not aligned with the target objects semantic boundaries.
    \item \textbf{Enclosed background regions as part of the segment:} Ground truth segments like chairs or bicycles often include enclosed background. Since these regions are semantically not similar to the target object, our model might struggle to segment these regions.
\end{itemize}

\section{Visualization of Score Functions}
The example in \cref{fig:score_metrics_graphs_simple} shows the segmentation of an image from the GrabCut dataset with three prompt points.
For each prompt point, we display the corresponding Markov-map, the score functions $s_{i, \cdot}(\lambda)$ evaluations of potential thresholds $\lambda$ and the resulting single point segmentation using the highest scoring threshold $\lambda_i = \underset{\lambda}{\arg \max}~ s_i(\lambda)$.
As introduced in \cref{eqn:total_score_function_definition}, the total score is a product of all four score functions $s_{i, \mathrm{prior}}(\lambda)$, $s_{i, \mathrm{edge}}(\lambda)$, $s_{i, \mathrm{pos}}(\lambda)$ and $s_{i, \mathrm{neg}}(\lambda)$.
Looking at the second prompt point in the middle column, we observe the edge score function $s_{i, \mathrm{edge}}(\lambda)$ shows two local maxima.
By multiplying it with the $s_{i, \mathrm{pos}}(\lambda)$ function values, the second local maximum increases and therefore the threshold containing the entire mushroom will be selected, as can be seen in the last row.
We additionally provide the segmentation of a more complex image in \cref{fig:score_metrics_graphs}.
The segmentation target is the person on the right (including the held bottle).

\section{Score Function Scenarios}
\cref{fig:score_metrics_graphs} shows segmentation examples of each score function.
Starting in the first columns on going from left to right:
\\
\\
\textbf{Column 1, No score function}: The threshold is constant $\lambda_i=0.5$.
We observe in the first row that the segmentation of the wolf does not align with the boundaries of the segment.
The second row predicts the wolf correctly due to the contribution of the $s_{edge}(\lambda)$ function.
\\
\\
\textbf{Column 2, No $s_{prior}$}: This score function ensures that the segmentation of each individual prompt point can not exceed a size of $40\%$ of the image.
Without it, in a few cases, we observe that the semantically strong difference between sky and ground results in M2N2 predicting the entire foreground as segment.
We therefore restrict the per point segment size to prevent such cases.
\\
\\
\textbf{Column 3, No $s_{edge}$}: The function contributes to higher scores for thresholds $\lambda$ being better aligned with the semantic boundary of the Markov-map. Without it, in the first row the flower's segmentation is misaligned with both the petal and the ovary.
\\
\\
\textbf{Column 4, No $s_{pos}$}: This score function provides higher scores for segments contain more points of the same class. The kangaroo in the first row is segmented with two disconnected regions, the first point segments the body and head, the second point segments the tip of the tail. In the second row, the $s_{pos}$ scores thresholds of which the resulting segment contains both points higher. Therefore we get a different threshold resulting in a connected segment with both points inside and the foot of the kangaroo included.
\\
\\
\textbf{Column 5, No $s_{neg}$}: The $s_{neg}$ ensures that no prompt points of the other class are within the segment.
The challenge of the segmentation in the first row is that the green foreground point at the bottom segments the foreground region in focus and the red background point segments the two zebras.
Since the two zebras on the left also belong to the lower foreground region, the foreground and background point's segments overlap.
This results in the segmentation having no clear boundary.
By using the $s_{neg}$ score function, as can be seen in the second row, the problem is resolved.
The contribution of $s_{neg}$ results in a smaller segment size of the lower green foreground point, putting the focus of the segmentation on the foot of the zebra of interest.

\clearpage
\begin{figure}
  \centering{\includegraphics[width=\columnwidth]{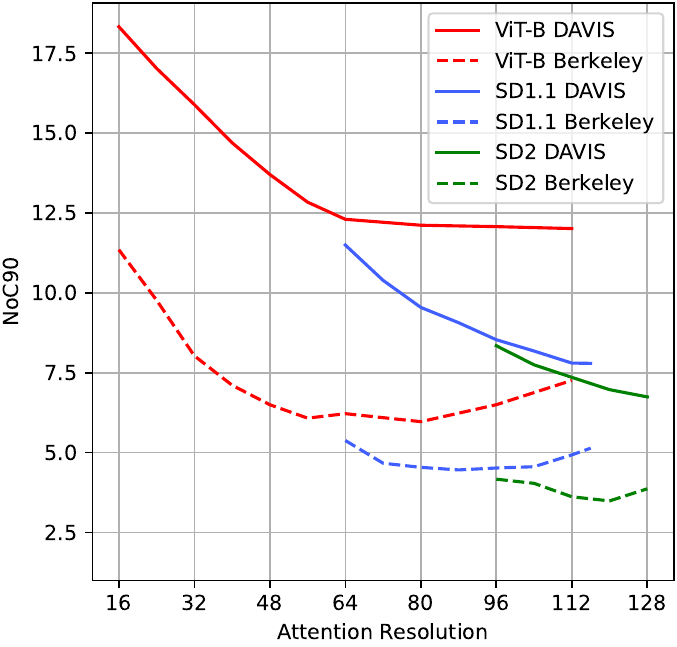}}
  \caption{\textbf{Impact of attention resolution on various backbones}. We evaluate attention maps starting from the native training resolution up to the highest possible resolution on our hardware.}
  \label{fig:backbone_comparison}
\end{figure}
\section{Additional Backbones in Detail} \label{sec:additional_backbones_in_detail}
\textbf{ViT-B}: We use the pre-trained weights of DinoV2 provided by Hugging Face's transformers package. During extraction of the attention maps, we remove the CLS token to obtain the image self-attention. We aggregate only the last attention layer as it results in the lowest NoC. This makes sense, as DinoV2's training objective is to obtain rich semantic features in the last layer of the ViT. Furthermore, DinoV2's training process involves two stages of which the first stage trains on an image resolution of $224 \times 224$ and the later stage does fine-tuning on the higher-res image resolution of $518 \times 518$.
With a patch size of $14 \times 14$ pixels, we obtain an attention tensor of the shape $16 \times 16  \times 16 \times 16$ for an image resolution of $224 \times 224$ and $37 \times 37  \times 37 \times 37$ for an image resolution of $518 \times 518$, respectively.
For our comparison in \cref{tab:main_comparison}, we choose an attention resolution of $80$ because higher resolutions increase the NoC as can be observed in \cref{fig:backbone_comparison}. All other hyperparameters are the same as in the SD2 backbone.\\
\\
\textbf{SD1.1}: As SD1.1 and SD2 use the same model architecture, we use the same hyperparameters for both models except the attention resolution which we set to $112$ for the SD1.1 model.

\section{Choice of Backbone} \label{sec:choice_of_backbone}
M2N2's NoC is highly dependent on the attention map resolution. \cref{fig:backbone_comparison} shows a comparison of various backbones on the DAVIS and Berkeley NoC90 values.
Each backbone is evaluated on a range of attention map resolutions, starting with the lowest resolution used during training.
We observe that increasing the attention resolution of ViT-B from the training resolution of $37$ to $64$ strongly improves the NoC.
Beyond a resolution of $64$, we note that ViT-B's NoC only decreases minimally for DAVIS and even increases for attention resolutions greater than $80$ for Berkeley.
This is probably due to the fact that an attention map resolution of $64 \times 64$ requires an input image resolution of $896 \times 896$, which is significantly higher than the largest image resolution of $518 \times 518$ used during training.
We observe something similar with SD1.1, where increasing the attention map resolution beyond $96$ starts to increase the NoC.
We therefore assume that increasing the attention resolution artificially by scaling up the input image resolution results in diminishing returns if the image resolution is significantly higher than the resolution used during training.
For this reason, we choose SD2 as the main backbone of our paper, as it is trained on an image resolution of $768 \times 768$ which results in the highest training attention map resolution of $96 \times 96$.

%% file: main.bbl
\begin{thebibliography}{54}
\providecommand{\natexlab}[1]{#1}
\providecommand{\url}[1]{\texttt{#1}}
\expandafter\ifx\csname urlstyle\endcsname\relax
  \providecommand{\doi}[1]{doi: #1}\else
  \providecommand{\doi}{doi: \begingroup \urlstyle{rm}\Url}\fi

\bibitem[Ambellan et~al.(2018)Ambellan, Tack, Ehlke, and Zachow]{OAIZIB}
Felix Ambellan, Alexander Tack, Moritz Ehlke, and Stefan Zachow.
\newblock Automated segmentation of knee bone and cartilage combining statistical shape knowledge and convolutional neural networks: Data from the osteoarthritis initiative.
\newblock In \emph{Medical Imaging with Deep Learning}, 2018.

\bibitem[Bai and Sapiro(2009)]{geodesicMattingSegmentation}
Xue Bai and Guillermo Sapiro.
\newblock Geodesic matting: A framework for fast interactive image and video segmentation and matting.
\newblock \emph{Int. J. Comput. Vision}, 82\penalty0 (2):\penalty0 113–132, 2009.

\bibitem[{Baid} et~al.(2021){Baid}, {Ghodasara}, {Mohan}, {Bilello}, {Calabrese}, {Colak}, et~al.]{BraTS}
Ujjwal {Baid}, Satyam {Ghodasara}, Suyash {Mohan}, Michel {Bilello}, Evan {Calabrese}, Errol {Colak}, et~al.
\newblock {The RSNA-ASNR-MICCAI BraTS 2021 Benchmark on Brain Tumor Segmentation and Radiogenomic Classification}.
\newblock \emph{arXiv e-prints}, art. arXiv:2107.02314, 2021.

\bibitem[Caron et~al.(2021)Caron, Touvron, Misra, J'egou, Mairal, Bojanowski, and Joulin]{DINO}
Mathilde Caron, Hugo Touvron, Ishan Misra, Herv'e J'egou, Julien Mairal, Piotr Bojanowski, and Armand Joulin.
\newblock Emerging properties in self-supervised vision transformers.
\newblock \emph{2021 IEEE/CVF International Conference on Computer Vision (ICCV)}, pages 9630--9640, 2021.

\bibitem[Chen et~al.(2023)Chen, Sun, Song, and Luo]{DiffusionDet}
Shoufa Chen, Pei Sun, Yibing Song, and Ping Luo.
\newblock Diffusiondet: Diffusion model for object detection.
\newblock In \emph{IEEE/CVF International Conference on Computer Vision (ICCV)}, 2023.

\bibitem[Chen et~al.(2022)Chen, Zhao, Zhang, Duan, Qi, and Zhao]{FocalClick}
Xi Chen, Zhiyan Zhao, Yilei Zhang, Manni Duan, Donglian Qi, and Hengshuang Zhao.
\newblock { FocalClick: Towards Practical Interactive Image Segmentation }.
\newblock In \emph{2022 IEEE/CVF Conference on Computer Vision and Pattern Recognition (CVPR)}, pages 1290--1299, Los Alamitos, CA, USA, 2022. IEEE Computer Society.

\bibitem[Dosovitskiy et~al.(2021)Dosovitskiy, Beyer, Kolesnikov, Weissenborn, Zhai, Unterthiner, Dehghani, Minderer, Heigold, Gelly, Uszkoreit, and Houlsby]{vit}
Alexey Dosovitskiy, Lucas Beyer, Alexander Kolesnikov, Dirk Weissenborn, Xiaohua Zhai, Thomas Unterthiner, Mostafa Dehghani, Matthias Minderer, Georg Heigold, Sylvain Gelly, Jakob Uszkoreit, and Neil Houlsby.
\newblock An image is worth 16x16 words: Transformers for image recognition at scale.
\newblock In \emph{9th International Conference on Learning Representations, {ICLR} 2021, Virtual Event, Austria, May 3-7, 2021}. OpenReview.net, 2021.

\bibitem[Grady(2006)]{randomWalkSegmentation}
Leo Grady.
\newblock Random walks for image segmentation.
\newblock \emph{IEEE Trans. Pattern Anal. Mach. Intell.}, 28\penalty0 (11):\penalty0 1768–1783, 2006.

\bibitem[Gulshan et~al.(2010)Gulshan, Rother, Criminisi, Blake, and Zisserman]{geodesicStartSegmentation}
Varun Gulshan, Carsten Rother, Antonio Criminisi, Andrew Blake, and Andrew Zisserman.
\newblock Geodesic star convexity for interactive image segmentation.
\newblock In \emph{2010 IEEE Computer Society Conference on Computer Vision and Pattern Recognition}, pages 3129--3136, 2010.

\bibitem[Gupta et~al.(2019)Gupta, Dollar, and Girshick]{LVIS}
Agrim Gupta, Piotr Dollar, and Ross Girshick.
\newblock Lvis: A dataset for large vocabulary instance segmentation.
\newblock In \emph{Proceedings of the IEEE/CVF Conference on Computer Vision and Pattern Recognition (CVPR)}, 2019.

\bibitem[Hamilton et~al.(2022)Hamilton, Zhang, Hariharan, Snavely, and Freeman]{hamilton2022unsupervised}
Mark Hamilton, Zhoutong Zhang, Bharath Hariharan, Noah Snavely, and William~T. Freeman.
\newblock Unsupervised semantic segmentation by distilling feature correspondences.
\newblock In \emph{International Conference on Learning Representations (ICLR)}, 2022.

\bibitem[Hariharan et~al.(2011)Hariharan, Bourdev, Arbelaez, Malik, and Maji]{SBD-Dataset}
Bharath Hariharan, Lubomir Bourdev, Pablo Arbelaez, Jitendra Malik, and Subhransu Maji.
\newblock { Semantic contours from inverse detectors }.
\newblock In \emph{2011 IEEE International Conference on Computer Vision (ICCV 2011)}, pages 991--998, Los Alamitos, CA, USA, 2011. IEEE Computer Society.

\bibitem[He et~al.(2021)He, Chen, Xie, Li, Doll'ar, and Girshick]{MAE}
Kaiming He, Xinlei Chen, Saining Xie, Yanghao Li, Piotr Doll'ar, and Ross~B. Girshick.
\newblock Masked autoencoders are scalable vision learners.
\newblock \emph{2022 IEEE/CVF Conference on Computer Vision and Pattern Recognition (CVPR)}, pages 15979--15988, 2021.

\bibitem[Huang et~al.(2023)Huang, Yang, Sun, Zhang, Cao, Jiang, and Ji]{interformer}
You Huang, Hao Yang, Ke Sun, Shengchuan Zhang, Liujuan Cao, Guannan Jiang, and Rongrong Ji.
\newblock Interformer real-time interactive image segmentation.
\newblock In \emph{ICCV}, pages 22244--22254. IEEE, 2023.

\bibitem[Jang and Kim(2019)]{interactiveSegmentationbackpropRefinement}
Won-Dong Jang and Chang-Su Kim.
\newblock Interactive image segmentation via backpropagating refinement scheme.
\newblock In \emph{2019 IEEE/CVF Conference on Computer Vision and Pattern Recognition (CVPR)}, pages 5292--5301, 2019.

\bibitem[Ke et~al.(2023{\natexlab{a}})Ke, Obukhov, Huang, Metzger, Daudt, and Schindler]{Repurposing_SD_for_MonoDepth}
Bingxin Ke, Anton Obukhov, Shengyu Huang, Nando Metzger, Rodrigo~Caye Daudt, and Konrad Schindler.
\newblock Repurposing diffusion-based image generators for monocular depth estimation.
\newblock \emph{2024 IEEE/CVF Conference on Computer Vision and Pattern Recognition (CVPR)}, pages 9492--9502, 2023{\natexlab{a}}.

\bibitem[Ke et~al.(2023{\natexlab{b}})Ke, Ye, Danelljan, liu, Tai, Tang, and Yu]{HQSAM}
Lei Ke, Mingqiao Ye, Martin Danelljan, Yifan liu, Yu-Wing Tai, Chi-Keung Tang, and Fisher Yu.
\newblock Segment anything in high quality.
\newblock In \emph{Advances in Neural Information Processing Systems}, pages 29914--29934. Curran Associates, Inc., 2023{\natexlab{b}}.

\bibitem[Kirillov et~al.(2023)Kirillov, Mintun, Ravi, Mao, Rolland, Gustafson, Xiao, Whitehead, Berg, Lo, Dollár, and Girshick]{SAM}
Alexander Kirillov, Eric Mintun, Nikhila Ravi, Hanzi Mao, Chloe Rolland, Laura Gustafson, Tete Xiao, Spencer Whitehead, Alexander~C. Berg, Wan-Yen Lo, Piotr Dollár, and Ross Girshick.
\newblock Segment anything.
\newblock In \emph{2023 IEEE/CVF International Conference on Computer Vision (ICCV)}, pages 3992--4003, 2023.

\bibitem[Kopf et~al.(2007)Kopf, Cohen, Lischinski, and Uyttendaele]{JBU}
Johannes Kopf, Michael~F. Cohen, Dani Lischinski, and Matthew Uyttendaele.
\newblock Joint bilateral upsampling.
\newblock \emph{ACM SIGGRAPH 2007 papers}, 2007.

\bibitem[Li et~al.(2023{\natexlab{a}})Li, Prabhudesai, Duggal, Brown, and Pathak]{SD_Secretly_Zero_Shot_Classifier}
Alexander~C. Li, Mihir Prabhudesai, Shivam Duggal, Ellis Brown, and Deepak Pathak.
\newblock { Your Diffusion Model is Secretly a Zero-Shot Classifier }.
\newblock In \emph{2023 IEEE/CVF International Conference on Computer Vision (ICCV)}, pages 2206--2217, Los Alamitos, CA, USA, 2023{\natexlab{a}}. IEEE Computer Society.

\bibitem[Li et~al.(2023{\natexlab{b}})Li, Zhao, Wang, Cheng, Jin, Ji, ming Yuan, Liu, and Chen]{MIS}
Kehan Li, Yian Zhao, Zhennan Wang, Zesen Cheng, Peng Jin, Xiang Ji, Li ming Yuan, Chang Liu, and Jie Chen.
\newblock Multi-granularity interaction simulation for unsupervised interactive segmentation.
\newblock \emph{2023 IEEE/CVF International Conference on Computer Vision (ICCV)}, pages 666--676, 2023{\natexlab{b}}.

\bibitem[Li et~al.(2024)Li, Lin, Chen, Liu, Wang, Singh, and Raj]{paintSeg}
Xiang Li, Chung-Ching Lin, Yinpeng Chen, Zicheng Liu, Jinglu Wang, Rita Singh, and Bhiksha Raj.
\newblock Paintseg: training-free segmentation via painting.
\newblock In \emph{Proceedings of the 37th International Conference on Neural Information Processing Systems}, Red Hook, NY, USA, 2024. Curran Associates Inc.

\bibitem[Lin et~al.(2014)Lin, Maire, Belongie, Bourdev, Girshick, Hays, Perona, Ramanan, Doll{'{a} }r, and Zitnick]{COCO}
Tsung{-}Yi Lin, Michael Maire, Serge~J. Belongie, Lubomir~D. Bourdev, Ross~B. Girshick, James Hays, Pietro Perona, Deva Ramanan, Piotr Doll{'{a} }r, and C.~Lawrence Zitnick.
\newblock Microsoft {COCO:} common objects in context.
\newblock \emph{CoRR}, abs/1405.0312, 2014.

\bibitem[Lin et~al.(2020)Lin, Zhang, Chen, Cheng, and Lu]{interactiveSegmentationFirstClickAttn}
Zheng Lin, Zhao Zhang, Lin-Zhuo Chen, Ming-Ming Cheng, and Shao-Ping Lu.
\newblock Interactive image segmentation with first click attention.
\newblock In \emph{2020 IEEE/CVF Conference on Computer Vision and Pattern Recognition (CVPR)}, pages 13336--13345, 2020.

\bibitem[Lin et~al.(2022)Lin, Duan, Zhang, Guo, and Cheng]{interactiveSegmentationFocusCut}
Zheng Lin, Zheng-Peng Duan, Zhao Zhang, Chun-Le Guo, and Ming-Ming Cheng.
\newblock Focuscut: Diving into a focus view in interactive segmentation.
\newblock In \emph{Proceedings of the IEEE/CVF Conference on Computer Vision and Pattern Recognition (CVPR)}, pages 2637--2646, 2022.

\bibitem[Liu et~al.(2024{\natexlab{a}})Liu, Wang, Yin, Sonke, and Gavves]{clickPromptOptimalTransport}
Jie Liu, Haochen Wang, Wenzhe Yin, Jan-Jakob Sonke, and Efstratios Gavves.
\newblock Click prompt learning with optimal transport for interactive segmentation.
\newblock In \emph{European Conference on Computer Vision (ECCV)}, 2024{\natexlab{a}}.

\bibitem[Liu et~al.(2023)Liu, Xu, Bertasius, and Niethammer]{SimpleClick}
Qin Liu, Zhenlin Xu, Gedas Bertasius, and Marc Niethammer.
\newblock Simpleclick: Interactive image segmentation with simple vision transformers.
\newblock In \emph{Proceedings. IEEE International Conference on Computer Vision}, pages 22233--22243, 2023.

\bibitem[Liu et~al.(2024{\natexlab{b}})Liu, Cho, Bansal, and Niethammer]{SegNext-InteractiveSegmentation}
Qin Liu, Jaemin Cho, Mohit Bansal, and Marc Niethammer.
\newblock Rethinking interactive image segmentation with low latency, high quality, and diverse prompts.
\newblock In \emph{{IEEE/CVF} Conference on Computer Vision and Pattern Recognition, {CVPR} 2024, Seattle, WA, USA, June 16-22, 2024}, pages 3773--3782. {IEEE}, 2024{\natexlab{b}}.

\bibitem[Long et~al.(2015)Long, Shelhamer, and Darrell]{FCNSemanticSegmentation}
Jonathan Long, Evan Shelhamer, and Trevor Darrell.
\newblock Fully convolutional networks for semantic segmentation.
\newblock In \emph{2015 IEEE Conference on Computer Vision and Pattern Recognition (CVPR)}, pages 3431--3440, 2015.

\bibitem[Martin et~al.(2001)Martin, Fowlkes, Tal, and Malik]{berkeleyDataset}
D. Martin, C. Fowlkes, D. Tal, and J. Malik.
\newblock A database of human segmented natural images and its application to evaluating segmentation algorithms and measuring ecological statistics.
\newblock In \emph{Computer Vision, 2001. ICCV 2001. Proceedings. Eighth IEEE International Conference on}, pages 416--423 vol.2, 2001.

\bibitem[Melas-Kyriazi et~al.(2022)Melas-Kyriazi, Rupprecht, Laina, and Vedaldi]{DSM}
Luke Melas-Kyriazi, Christian Rupprecht, Iro Laina, and Andrea Vedaldi.
\newblock Deep spectral methods: A surprisingly strong baseline for unsupervised semantic segmentation and localization.
\newblock In \emph{2022 IEEE Conference on Computer Vision and Pattern Recognition (CVPR)}, 2022.

\bibitem[Oquab et~al.(2024)Oquab, Darcet, Moutakanni, Vo, Szafraniec, Khalidov, Fernandez, HAZIZA, Massa, El-Nouby, Assran, Ballas, Galuba, Howes, Huang, Li, Misra, Rabbat, Sharma, Synnaeve, Xu, Jegou, Mairal, Labatut, Joulin, and Bojanowski]{DINOv2}
Maxime Oquab, Timoth{\'e}e Darcet, Th{\'e}o Moutakanni, Huy~V. Vo, Marc Szafraniec, Vasil Khalidov, Pierre Fernandez, Daniel HAZIZA, Francisco Massa, Alaaeldin El-Nouby, Mido Assran, Nicolas Ballas, Wojciech Galuba, Russell Howes, Po-Yao Huang, Shang-Wen Li, Ishan Misra, Michael Rabbat, Vasu Sharma, Gabriel Synnaeve, Hu Xu, Herve Jegou, Julien Mairal, Patrick Labatut, Armand Joulin, and Piotr Bojanowski.
\newblock {DINO}v2: Learning robust visual features without supervision.
\newblock \emph{Transactions on Machine Learning Research}, 2024.

\bibitem[Perazzi et~al.(2016)Perazzi, Pont-Tuset, McWilliams, Van~Gool, Gross, and Sorkine-Hornung]{DAVIS_DAtaset}
Federico Perazzi, Jordi Pont-Tuset, Brian McWilliams, Luc Van~Gool, Markus Gross, and Alexander Sorkine-Hornung.
\newblock A benchmark dataset and evaluation methodology for video object segmentation.
\newblock In \emph{Proceedings of the IEEE Conference on Computer Vision and Pattern Recognition (CVPR)}, 2016.

\bibitem[Rana et~al.(2023)Rana, Mahadevan, Hermans, and Leibe]{Rana:Dynamite}
Amit~Kumar Rana, Sabarinath Mahadevan, Alexander Hermans, and Bastian Leibe.
\newblock {D}yna{MIT}e : {D}ynamic {Q}uery {B}ootstrapping for {M}ulti-object {I}nteractive {S}egmentation {T}ransformer.
\newblock In \emph{2023 IEEE/CVF International Conference on Computer Vision workshops : ICCVW 2023 : Paris, France, 2-6 October 2023 : proceedings / general chairs: Jana Kosecka (GMU), Jean Ponce (ENS-PSl and NYU), Cordelia Schmid (Inria/Google), Andrew Zisserman (Oxford/Deepmind) ; publications chairs: Frédéric Jurie (ENSICAEN), Gaurav Sharma (TensorTour and IIT Kanpur) ; publisher: IEEE}, pages 1043--1052, Piscataway, NJ, 2023. 2023 IEEE/CVF International Conference on Computer Vision, Paris (France), 1 Oct 2023 - 6 Oct 2023, IEEE.

\bibitem[Ravi et~al.(2024)Ravi, Gabeur, Hu, Hu, Ryali, Ma, Khedr, Rädle, Rolland, Gustafson, Mintun, Pan, Alwala, Carion, Wu, Girshick, Dollár, and Feichtenhofer]{SAM2}
Nikhila Ravi, Valentin Gabeur, Yuan-Ting Hu, Ronghang Hu, Chaitanya Ryali, Tengyu Ma, Haitham Khedr, Roman Rädle, Chloe Rolland, Laura Gustafson, Eric Mintun, Junting Pan, Kalyan~Vasudev Alwala, Nicolas Carion, Chao-Yuan Wu, Ross Girshick, Piotr Dollár, and Christoph Feichtenhofer.
\newblock Sam 2: Segment anything in images and videos, 2024.

\bibitem[Rombach et~al.(2022)Rombach, Blattmann, Lorenz, Esser, and Ommer]{StableDiffusion}
Robin Rombach, Andreas Blattmann, Dominik Lorenz, Patrick Esser, and Bj{\"o}rn Ommer.
\newblock High-resolution image synthesis with latent diffusion models.
\newblock In \emph{Proceedings of the IEEE/CVF Conference on Computer Vision and Pattern Recognition}, pages 10684--10695, 2022.

\bibitem[Ross(1997)]{Ross97}
Sheldon~M. Ross.
\newblock \emph{Introduction to Probability Models}.
\newblock Academic Press, San Diego, CA, USA, sixth edition, 1997.

\bibitem[Rother et~al.(2004)Rother, Kolmogorov, and Blake]{grabcut}
Carsten Rother, Vladimir Kolmogorov, and Andrew Blake.
\newblock "grabcut": interactive foreground extraction using iterated graph cuts.
\newblock In \emph{ACM SIGGRAPH 2004 Papers}, page 309–314, New York, NY, USA, 2004. Association for Computing Machinery.

\bibitem[Shi and Malik(2000)]{Shi00normalizedCuts}
Jianbo Shi and J. Malik.
\newblock Normalized cuts and image segmentation.
\newblock \emph{IEEE Trans. on Pattern Analysis and Machine Intelligence}, 22\penalty0 (8):\penalty0 888--905, 2000.

\bibitem[Sim{\'e}oni et~al.(2021)Sim{\'e}oni, Puy, Vo, Roburin, Gidaris, Bursuc, P'erez, Marlet, and Ponce]{Simoni2021LocalizingOW}
Oriane Sim{\'e}oni, Gilles Puy, Huy~V. Vo, Simon Roburin, Spyros Gidaris, Andrei Bursuc, Patrick P'erez, Renaud Marlet, and Jean Ponce.
\newblock Localizing objects with self-supervised transformers and no labels.
\newblock \emph{ArXiv}, abs/2109.14279, 2021.

\bibitem[Sinkhorn(1964)]{sinkhornDoublyStochastic}
Richard Sinkhorn.
\newblock {A Relationship Between Arbitrary Positive Matrices and Doubly Stochastic Matrices}.
\newblock \emph{The Annals of Mathematical Statistics}, 35\penalty0 (2):\penalty0 876 -- 879, 1964.

\bibitem[Sofiiuk et~al.(2020)Sofiiuk, Petrov, Barinova, and Konushin]{interactiveSegmentationfBRS}
Konstantin Sofiiuk, Ilya~A. Petrov, Olga Barinova, and Anton Konushin.
\newblock F-brs: Rethinking backpropagating refinement for interactive segmentation.
\newblock In \emph{{IEEE} Conference on Computer Vision and Pattern Recognition (CVPR)}. {IEEE}, 2020.

\bibitem[Sofiiuk et~al.(2022)Sofiiuk, Petrov, and Konushin]{ritm2022}
Konstantin Sofiiuk, Ilya~A Petrov, and Anton Konushin.
\newblock Reviving iterative training with mask guidance for interactive segmentation.
\newblock In \emph{2022 IEEE International Conference on Image Processing (ICIP)}, pages 3141--3145. IEEE, 2022.

\bibitem[Sun et~al.(2024)Sun, Xian, Xu, Capriotti, and Yao]{CFR-ICL}
Shoukun Sun, Min Xian, Fei Xu, Luca Capriotti, and Tiankai Yao.
\newblock Cfr-icl: Cascade-forward refinement with iterative click loss for interactive image segmentation.
\newblock In \emph{Technical Tracks 14}, number~5 in Proceedings of the AAAI Conference on Artificial Intelligence, pages 5017--5024. Association for the Advancement of Artificial Intelligence, 2024.
\newblock Publisher Copyright: Copyright {\textcopyright} 2024, Association for the Advancement of Artificial Intelligence.; 38th AAAI Conference on Artificial Intelligence, AAAI 2024 ; Conference date: 20-02-2024 Through 27-02-2024.

\bibitem[Tian et~al.(2023)Tian, Aggarwal, Colaco, Kira, and Gonz{\'a}lez-Franco]{DiffuseAttendSegment}
Junjiao Tian, Lavisha Aggarwal, Andrea Colaco, Zsolt Kira, and Mar Gonz{\'a}lez-Franco.
\newblock Diffuse, attend, and segment: Unsupervised zero-shot segmentation using stable diffusion.
\newblock \emph{2024 IEEE/CVF Conference on Computer Vision and Pattern Recognition (CVPR)}, pages 3554--3563, 2023.

\bibitem[Wang et~al.(2024{\natexlab{a}})Wang, Choudhuri, Zheng, Gao, Planche, Deng, Liu, Chen, Bagci, and Wu]{OISegmentation}
Bin Wang, Anwesa Choudhuri, Meng Zheng, Zhongpai Gao, Benjamin Planche, Andong Deng, Qin Liu, Terrence Chen, Ulas Bagci, and Ziyan Wu.
\newblock Order-aware interactive segmentation, 2024{\natexlab{a}}.

\bibitem[Wang et~al.(2021)Wang, Zhang, Shen, Kong, and Li]{DenseCL}
Xinlong Wang, Rufeng Zhang, Chunhua Shen, Tao Kong, and Lei Li.
\newblock Dense contrastive learning for self-supervised visual pre-training.
\newblock In \emph{Proc. IEEE Conf. Computer Vision and Pattern Recognition (CVPR)}, 2021.

\bibitem[Wang et~al.(2022{\natexlab{a}})Wang, Yu, De~Mello, Kautz, Anandkumar, Shen, and Alvarez]{FreeSOLO}
Xinlong Wang, Zhiding Yu, Shalini De~Mello, Jan Kautz, Anima Anandkumar, Chunhua Shen, and Jose~M. Alvarez.
\newblock Freesolo: Learning to segment objects without annotations.
\newblock In \emph{Proceedings of the IEEE/CVF Conference on Computer Vision and Pattern Recognition (CVPR)}, pages 14176--14186, 2022{\natexlab{a}}.

\bibitem[Wang et~al.(2024{\natexlab{b}})Wang, Yang, and Darrell]{wang2024segment}
Xudong Wang, Jingfeng Yang, and Trevor Darrell.
\newblock Segment anything without supervision.
\newblock In \emph{The Thirty-eighth Annual Conference on Neural Information Processing Systems}, 2024{\natexlab{b}}.

\bibitem[Wang et~al.(2022{\natexlab{b}})Wang, Shen, Hu, Yuan, Crowley, and Vaufreydaz]{Wang2022SelfSupervisedTF}
Yangtao Wang, XI Shen, Shell~Xu Hu, Yuan Yuan, James~L. Crowley, and Dominique Vaufreydaz.
\newblock Self-supervised transformers for unsupervised object discovery using normalized cut.
\newblock \emph{2022 IEEE/CVF Conference on Computer Vision and Pattern Recognition (CVPR)}, pages 14523--14533, 2022{\natexlab{b}}.

\bibitem[Wang et~al.(2022{\natexlab{c}})Wang, Shen, Yuan, Du, Li, Hu, Crowley, and Vaufreydaz]{TokenCutSO}
Yangtao Wang, Xiaoke Shen, Yuan Yuan, Yuming Du, Maomao Li, Shell~Xu Hu, James~L. Crowley, and Dominique Vaufreydaz.
\newblock Tokencut: Segmenting objects in images and videos with self-supervised transformer and normalized cut.
\newblock \emph{IEEE Transactions on Pattern Analysis and Machine Intelligence}, 45:\penalty0 15790--15801, 2022{\natexlab{c}}.

\bibitem[Xu et~al.(2016)Xu, Price, Cohen, Yang, and Huang]{DIOS}
Ning Xu, Brian Price, Scott Cohen, Jimei Yang, and Thomas Huang.
\newblock Deep interactive object selection.
\newblock In \emph{2016 IEEE Conference on Computer Vision and Pattern Recognition (CVPR)}, pages 373--381, 2016.

\bibitem[Yang et~al.(2024)Yang, Xu, Kang, Shi, and Zhao]{FreeMask}
Lihe Yang, Xiaogang Xu, Bingyi Kang, Yinghuan Shi, and Hengshuang Zhao.
\newblock Freemask: synthetic images with dense annotations make stronger segmentation models.
\newblock In \emph{Proceedings of the 37th International Conference on Neural Information Processing Systems}, Red Hook, NY, USA, 2024. Curran Associates Inc.

\bibitem[Zhou et~al.(2023)Zhou, Wang, Zhao, Li, Huang, Meng, and Zheng]{interactiveSegmentationGaussianProcess}
Minghao Zhou, Hong Wang, Qian Zhao, Yuexiang Li, Yawen Huang, Deyu Meng, and Yefeng Zheng.
\newblock Interactive segmentation as gaussian process classification.
\newblock \emph{2023 IEEE/CVF Conference on Computer Vision and Pattern Recognition (CVPR)}, pages 19488--19497, 2023.

\end{thebibliography}
